\documentclass{article}

\usepackage[final]{neurips_2021}

\makeatletter
\renewcommand{\@noticestring}{%
      1st Workshop on Human and Machine Decisions
      (WHMD 2021) at NeurIPS 2021.}
\makeatother

\usepackage[utf8]{inputenc} % allow utf-8 input
\usepackage[T1]{fontenc}    % use 8-bit T1 fonts
\usepackage{hyperref}       % hyperlinks
\usepackage{url}            % simple URL typesetting
\usepackage{booktabs}       % professional-quality tables
\usepackage{amsfonts}       % blackboard math symbols
\usepackage{nicefrac}       % compact symbols for 1/2, etc.
\usepackage{microtype}      % microtypography
\usepackage{xcolor}         % colors

\usepackage{xspace,xcolor,bm,amsmath,graphicx,subcaption}

\newcommand\hull{$\mathcal{H}^{tr}$\xspace}

\makeatletter
\newcommand{\printfnsymbol}[1]{
  \textsuperscript{\@fnsymbol{#1}}
}

\title{Extrapolation Frameworks in Cognitive Psychology Suitable for Study of Image Classification Models}

\author{%
  Roozbeh Yousefzadeh\\%\thanks{Use footnote for providing further information
  Yale Center for Medical Informatics, Yale University\\ and VA Connecticut Healthcare System\\
  New Haven, CT 06510 \\
  \texttt{roozbeh.yousefzadeh@yale.edu} \\
    \AND
  Jessica A. Mollick\\
   Department of Psychiatry, Yale School of Medicine \\
   New Haven, CT 06510 \\
   \texttt{jessica.mollick@yale.edu} \\
}

\begin{document}

\maketitle

% \ry{Have been modifying the title several times (some of the variations are commented out above.). Feel free to change it entirely. It would be great if it appeals to people like Griffiths. The program committee is mostly computer scientists, so it would be great if it appeals to them as well, and also relates to the topic of workshop}

% \ry{This will go for a double blind review, so we have to hide our names, and in the text, we shouldn't reveal our identity.}

% \ry{Sent a message to organizers of the workshop on CMT to see if they would accept a 5 page submission}

\vspace{-.4cm}

\begin{abstract}
  We study the functional task of deep learning image classification models and show that image classification requires extrapolation capabilities. This suggests that new theories have to be developed for the understanding of deep learning as the current theory assumes models are solely interpolating, leaving many questions about them unanswered. We investigate the pixel space and also the feature spaces extracted from images by trained models (in their hidden layers, including the 64-dimensional feature space in the last hidden layer of pre-trained residual neural networks), and also the feature space extracted by wavelets/shearlets. In all these domains, testing samples considerably fall outside the convex hull of training sets, and image classification requires extrapolation. In contrast to the deep learning literature, in cognitive science, psychology, and neuroscience, extrapolation and learning are often studied in tandem. Moreover, many aspects of human visual cognition and behavior are reported to involve extrapolation. We propose a novel extrapolation framework for the mathematical study of deep learning models. In our framework, we use the term extrapolation in this specific way of extrapolating outside the convex hull of training set (in the pixel space or feature space) but within the specific scope defined by the training data, the same way extrapolation is defined in many studies in cognitive science. We explain that our extrapolation framework can provide novel answers to open research problems about deep learning including their over-parameterization, their training regime, out-of-distribution detection, etc. We also see that the extent of extrapolation is negligible in learning tasks where deep learning is reported to have no advantage over simple models.
\end{abstract}
% \jm{I suggest removing and/or defining the term "double descent" in here, I was unfamiliar. Otherwise looks good}\ry{Sure. fixed!}

% Technically, extrapolation can imply any classification that a model does outside the convex hull of training set even when the testing image is unrelated to the scope of training set. But, we expect to use a trained model on the specific scope that it is trained on. Therefore, in our framework, we use the term extrapolation in this specific way of extrapolating outside the convex hull of training set (in the pixel space or some feature space) but within the specific scope defined by the training data, the same way that extrapolation is defined in many studies in cognitive science.

\vspace{-.2cm}

\section{Introduction}

\vspace{-.2cm}

We propose an extrapolation framework to study the functional task of deep neural networks used for image classification. We first demonstrate that the functional task of image classification models involves extrapolation. Although extrapolation is not considered to explain the generalization of deep learning models, there is a rich literature in cognitive science, psychology, and neuroscience, studying different extrapolation tasks in relation to learning. We study that literature to see how extrapolation can be defined in different ways depending on the learning task. The definition we adopt in our extrapolation framework is \textbf{extrapolation within a specific scope defined by the training set}.

From the mathematical perspective, an image classification model can be considered a classification function that maps images to classes \citep{strang2019linear}. The domain of such function is the pixel space which can be considered a hyper-cube. The trained model partitions its domain and assigns a class to each partition. Partitions are defined by decision boundaries, and so is the model \citep{yousefzadeh2020deep}. This domain partitioning is inherent to the classification task of these models, and it can be studied not only in the pixel space but also in feature spaces derived from images in the internal layers of deep networks. Domain partitioning, indeed, has a rich literature in applied mathematics and approximation theory \citep{weierstrass1885analytische,fornaess2020holomorphic}. The training process of a model can be viewed as defining the partitions and decision boundaries. The location of decision boundaries inside the \underline{convex hull of training set (denoted by \hull)} can be viewed in relation to training samples, but geometry of data shows that testing samples can be considerably outside the \hull. Therefore, extensions of decision boundaries outside the \hull are crucial in model's generalization. This novel perspective can provide new answers to some open research questions in deep learning.

\vspace{-.2cm}

\section{Geometry of image classification datasets in pixel and in feature space}

\vspace{-.2cm}

% \ry{one option could be to make this section very very brief, move its content to the appendix, and refer the reader to appendix. Mostly likely, reviewers will not look at the appendix though. But this strategy could help with reducing our content to 4 pages. Open to other ideas as well.}

In this section, we provide a brief summary (from our previous work \citep{yousefzadeh2021hull}) about the geometry of image classification datasets (in the pixel and feature space). We report that there is no evidence that image classification is merely an interpolation task. Rather, all evidence indicate that the functional task of image classification models involve extrapolation. Studies on generalization of DNNs, however, are solely focused on interpolation, e.g., \citep{belkin2021fit,dar2021farewell,belkin2019does,belkin2019reconciling,ma2018power,belkin2018overfitting}, and extrapolation is completely absent from their framework of study.\footnote{See Appendix~\ref{sec:appx_interextra} for a discussion about definition of interpolation and extrapolation.} There are insightful studies on geometrical aspects of image classification models, e.g., \cite{cohen2020separability}, however, such studies do not consider extrapolation.

\begin{figure}[h]
\begin{center}
    \begin{minipage}[b]{.85\columnwidth}
    \centering
    (original) \raisebox{-.5\height}{\includegraphics[width=.12\columnwidth]{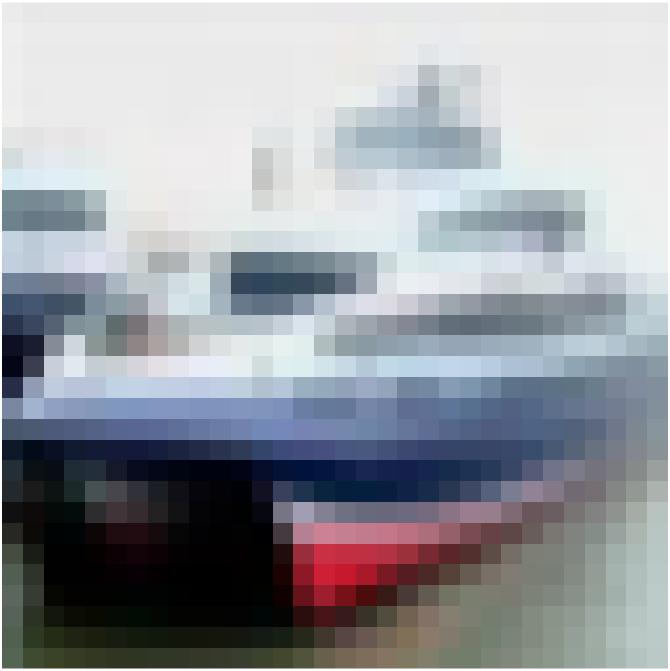}} {\vspace{.2cm} -}
    \raisebox{-.5\height}	{\includegraphics[width=.12\columnwidth]{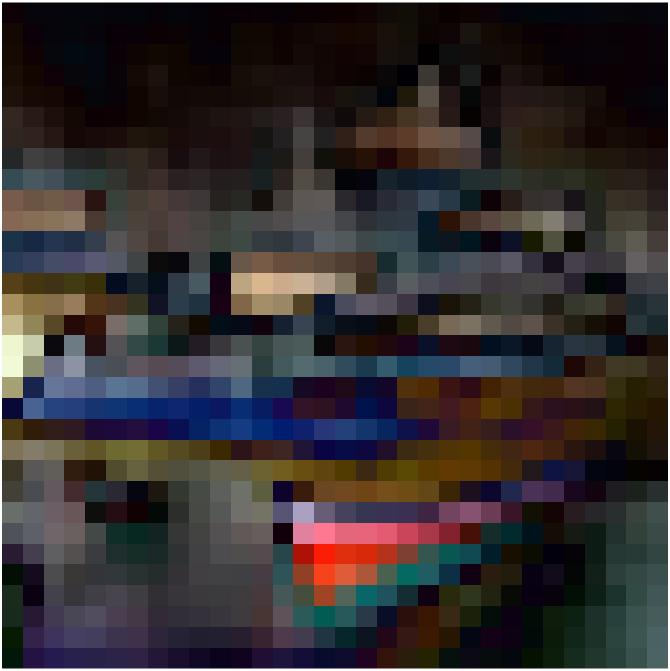}} =
    \raisebox{-.5\height}	{\includegraphics[width=.12\columnwidth]{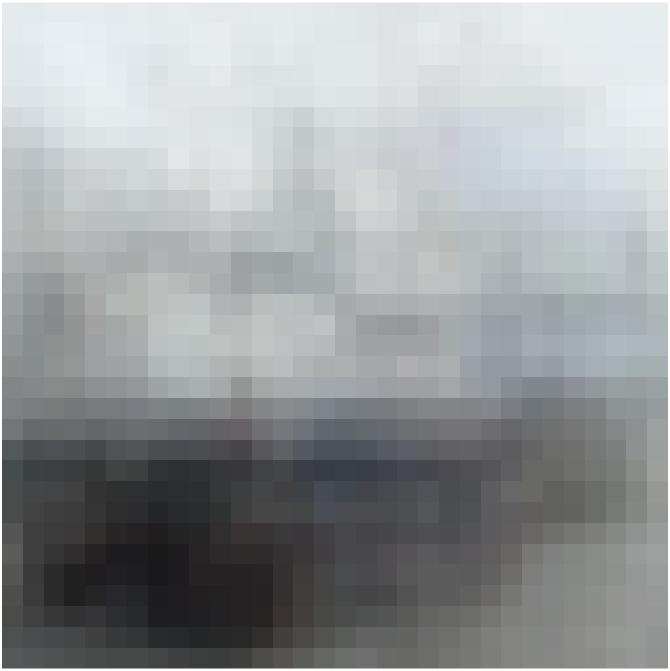}} (on \hull)
    \end{minipage}
    % \begin{minipage}[b]{.85\columnwidth}
    % \centering
    % (original) \raisebox{-.5\height}{\includegraphics[width=.12\columnwidth]{figures/cif10_te3956orig.jpg}} {\vspace{.2cm} -}
    % \raisebox{-.5\height}	{\includegraphics[width=.12\columnwidth]{figures/cif10_te3956diff_all.jpg}} =
    % \raisebox{-.5\height}	{\includegraphics[width=.12\columnwidth]{figures/cif10_te3956oncvx_all.jpg}} (on \hull)
    % \end{minipage}
\end{center}
    \caption{Minimum perturbation that can bring a testing image to \hull of all classes. (left image) original testing image from CIFAR-10, (middle image) what should be changed in the original image, (right image) the resulting image on the \hull. See Appendix \ref{sec:appx_projections} for more examples.}
  \label{fig:direction_to_hull}
\end{figure}

{\bf Extrapolation is significant even in 64-dimensional feature space learned by DNNs.} We consider the pixel space, the feature spaces that trained models derive from images throughout their hidden layers (including the 64-dimensional space in the last hidden layer of pre-trained residual neural networks.\footnote{Pre-trained model is available at \url{https://www.mathworks.com/help/deeplearning/ug/train-residual-network-for-image-classification.html}. Model has a residual network architecture \citep{he2016deep} with total depth of 20 layers.}), and also the feature spaces that one can derive from images using wavelets/shearlets.\footnote{For feature selection with wavelets and shearlets, we convolve the images with a wavelet basis such as Daubechies wavelets \citep{daubechies1992ten} or with a shearlet system \citep{andrade2020shearlets}. We combine the resulting coefficients in a matrix and then perform feature selection. For feature selection, we use rank-revealing QR factorization \citep{chan1987rank} which selects the features based on their rank contribution, and also use Laplacian score \citep{he2005laplacian} which selects features based on a nearest neighbor similarity graph. We arrive at similar distance distributions with respect to the resulting \hull's.} In all these domains, all testing samples of standard datasets such as MNIST, CIFAR-10, CIFAR-100, and some medical image datasets considerably fall outside their corresponding \hull. Moreover, distance of testing samples to \hull is considerable and each image has to visually change in order to reach the closest point on the surface of \hull, as shown in Figure~\ref{fig:direction_to_hull}. The direction that a model has to extrapolate outside its \hull, to reach and classify a testing sample, relate to important features in images and the objects of interest in them, coinciding with cognitive studies that suggest same visual regions in images are informative for both humans and machines \citep{langlois2021passive}. These visual features also signify that extrapolation task of a model is meaningful. This implies that the features that a model extracts and learns from images do not transform the functional task of the model into interpolation. In other words, testing images (that a trained model encounters) have some novelty that put them outside the \hull. For few-shot learning, extrapolation is even more significant, yet possible.

 %The number of features we select can be considered low-dimensional, e.g., for MNIST, all the trends remain similar when we select 50, 64, 100, 200, and 400 features from images (while the pixel space has 784 dimensions).

% For testing samples that are closest to the \hull, the original image and its projection to the \hull visually look similar to each other. On the other hand,

Projection of most testing samples to the \hull usually appear vague and ambiguous, and it would be hard (if not impossible) for a human to classify those projections without seeing the original image. Moreover, the trained model itself is not able to correctly classify the projections either. For example, if we use the projections of testing set of CIFAR-10 to the convex hull, instead of the original testing set, accuracy of a trained CNN would drop from above 90\% to 33\%. This also relates to confidence in correctness of classifications which is an open research problem in deep learning and psychology. While the psychological features learned from images are unknown, \cite{battleday2020capturing} studied how different psychological models fit human categorization behavior in the same dataset. Such psychological studies can be extended to the convex hull projection of images as well to provide insights about the extrapolation aspects of image classification.

% \jm{Suggest this rephrasing...While the psychological features learned from images are unknown, \cite{battleday2020capturing} studied how different psychological models fit human categorization behavior in the same dataset}\ry{fixed! this is definitely better}

\begin{figure}[h]
    \centering
     \begin{subfigure}[b]{0.48\textwidth}
         \centering
         \includegraphics[width=.18\linewidth]{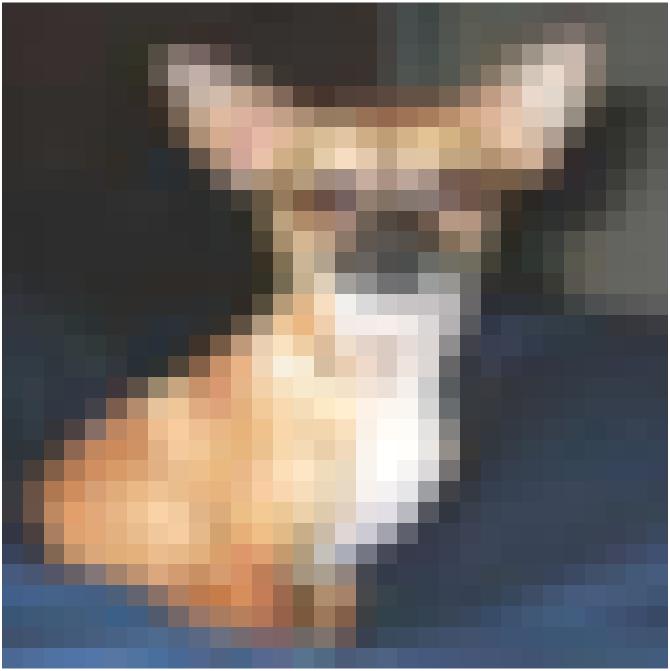}
         \includegraphics[width=.18\linewidth]{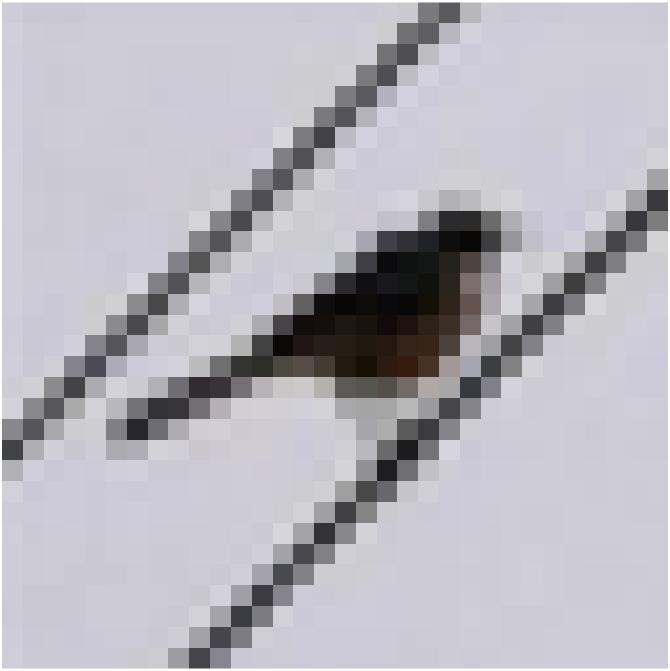}
         \includegraphics[width=.18\linewidth]{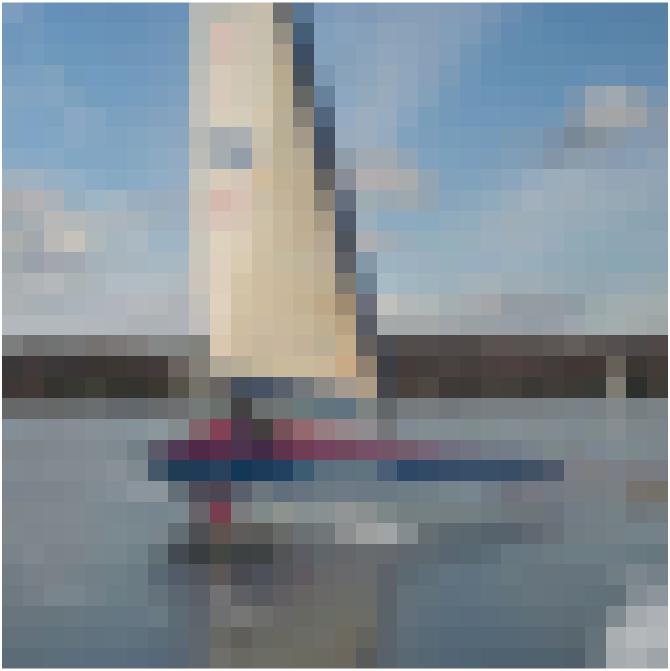}
         \includegraphics[width=.18\linewidth]{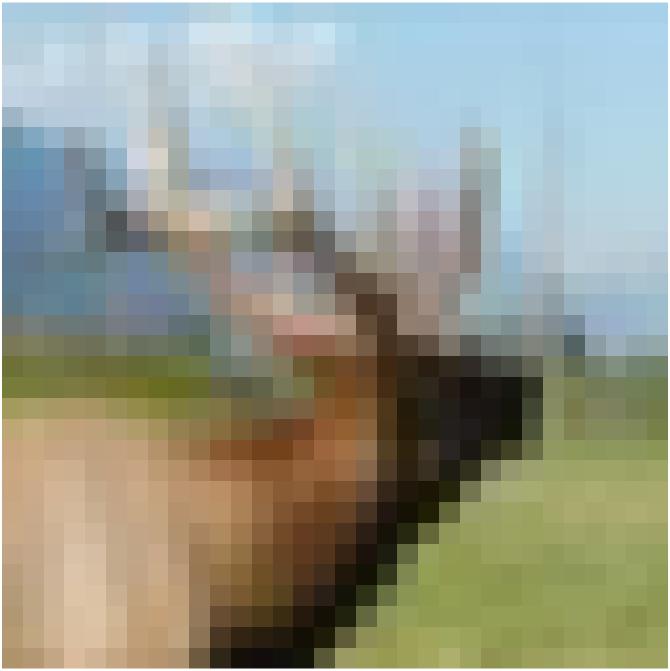}
         \includegraphics[width=.18\linewidth]{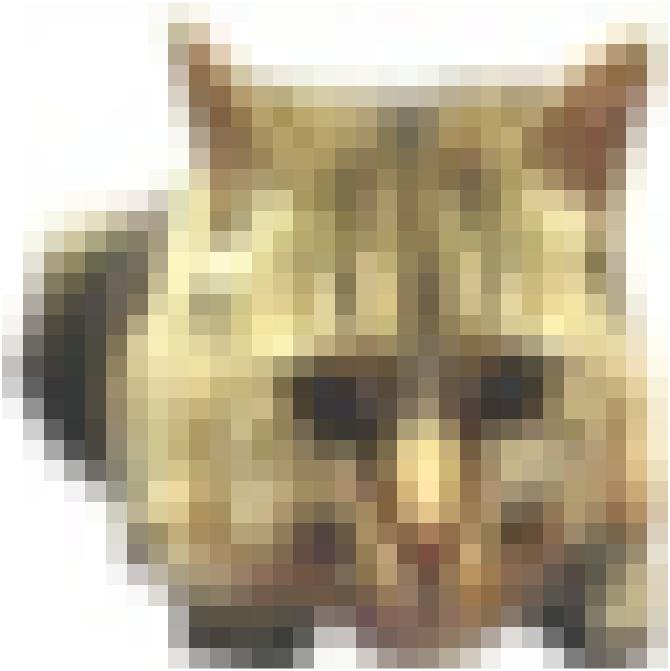}
         \caption{In feature space}
         \label{fig:far_mnist}
     \end{subfigure}
     \quad
     \begin{subfigure}[b]{0.48\textwidth}
         \centering
         \includegraphics[width=.18\linewidth]{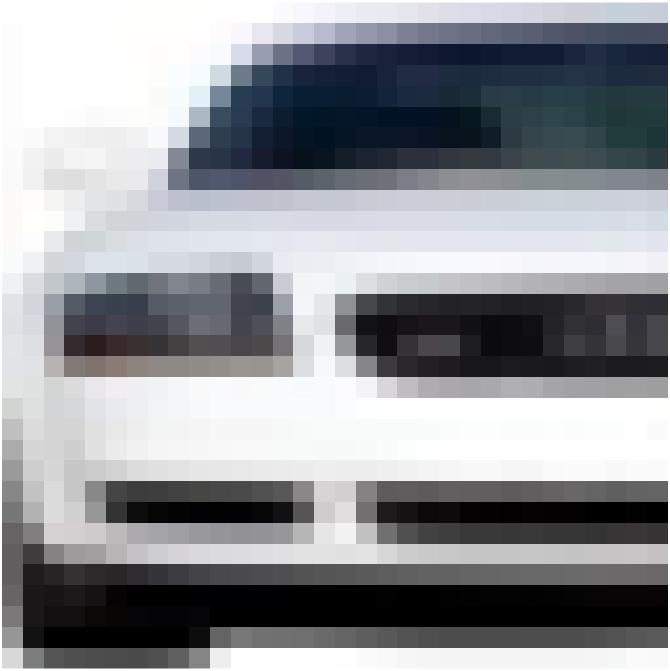}
         \includegraphics[width=.18\linewidth]{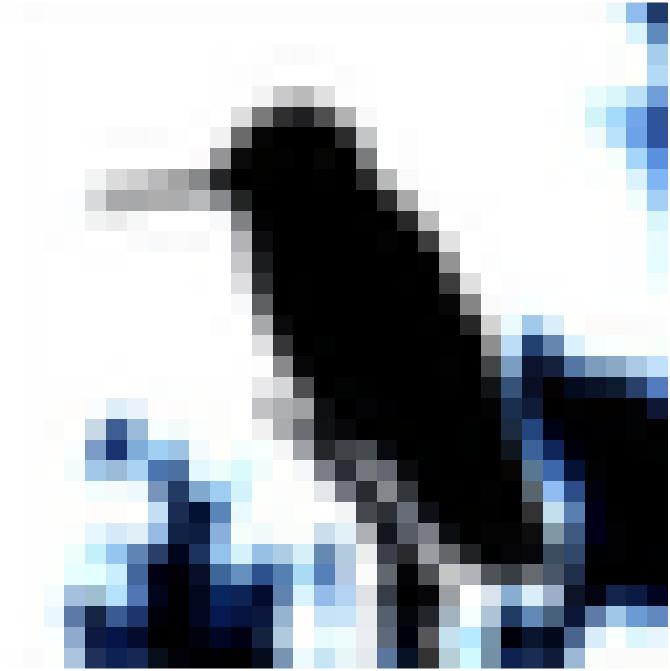}
         \includegraphics[width=.18\linewidth]{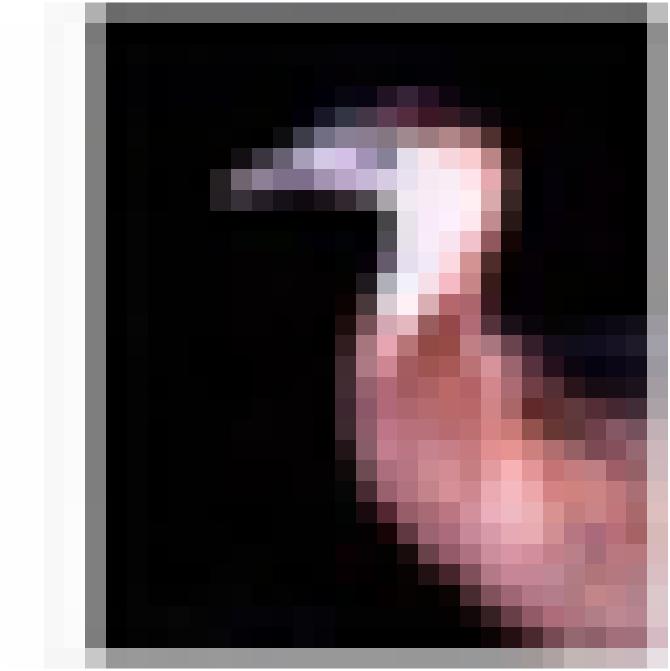}
         \includegraphics[width=.18\linewidth]{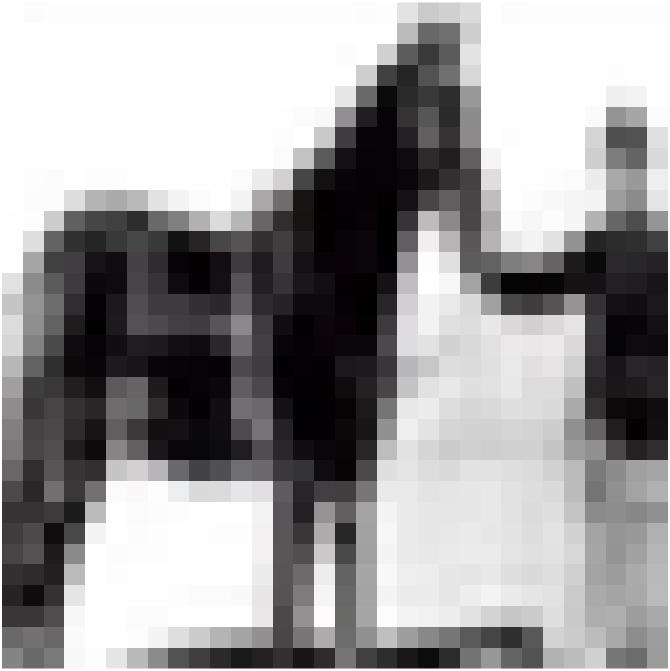}
         \includegraphics[width=.18\linewidth]{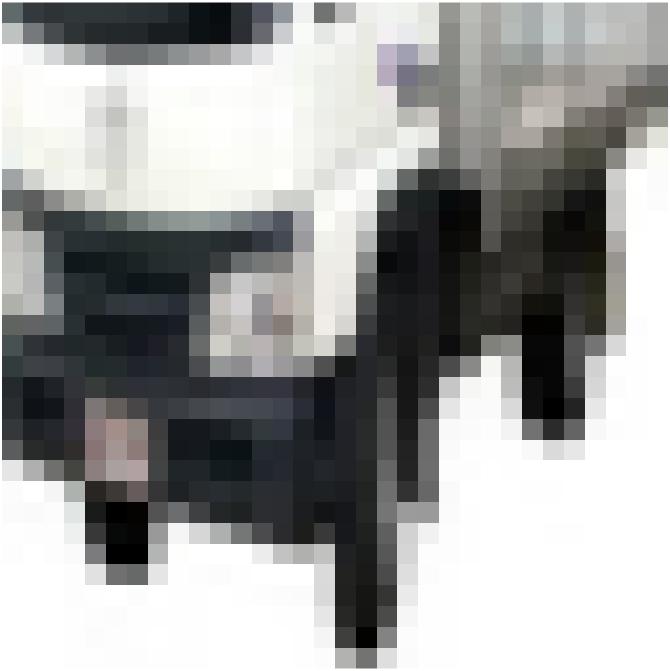}
         \caption{In pixel space}
         \label{fig:far_cif10}
     \end{subfigure}
    \caption{Samples of testing images from CIFAR-10 that are farthest from \hull. Feature space is the last hidden layer of a standard pre-trained CNN.}
  \label{fig:far_images}
\end{figure}

{\bf \hull of feature space differs from pixel space in meaningful ways.} %Please see Appendix~\ref{sec:appx_featurespace}.
% \section{\hull of feature space differs from pixel space in meaningful ways} \label{sec:appx_featurespace}
The feature spaces (in the last hidden layer of DNNs, and also derived by wavelets/shearlets) are meaningful and insightful compared to the pixel space. For example, testing samples that are far from the \hull in the feature space may be considered images with lack of similarity to samples in the training set. However, testing samples that are far from the \hull in the pixel space are predominantly white or black as shown in Figure~\ref{fig:far_images}. White and black colors are the extremes in the pixel space leading to larger distances. This distinction between the pixel space and feature space may correspond to the distinction between spatial learning and conceptual learning for non-image applications in cognitive science \citep{wu2020similarities}.
% \ry{sounds clear?} \jm{I'm not sure exactly, depends if Wu was talking about spatial learning specifically as applied to features of images or more generally (ie for navigation). Low-level visual features vs. conceptual learning?} \ry{I clarified that it is for non-image applications.}

{\bf Distance of testing images from the \hull is not too large.} Although all testing samples are outside the \hull and their distance is meaningful, the extent of extrapolation is relatively modest compared to the diameter of \hull. For example, for CIFAR-10, distance of testing samples varies between 1\% and 27\% of the diameter of \hull. This is noteworthy, because \hull may occupy a small fraction of our hyper-cube domain, and once we step outside the \hull, we may encounter many different types of images unrelated to the scope of training set. For example, outside the \hull of CIFAR-10, we may encounter radiology images, handwritten digits, etc. But, we do not expect a model trained on CIFAR-10 to classify radiology images. As a matter of definition, we intend to use a trained model only to classify images related to the scope of its training set. This motivates us to similarly limit our definition of extrapolation to the scope defined by the training set. In the next section, we see how extrapolation can be studied within a specific scope.

% While the closest point on the boundary of \hull is visually related to the query point, the closest training sample to the query point is usually unrelated and the direction between such points does not typically relate to relevant features in images.

% \section{Extrapolation within a specific scope: Taking inspiration from cognitive science and psychology}

% \ry{alternatively subsection title below: (It would be great if people like Griffiths find this section title and our paper title appealing)}

\vspace{-.2cm}

\section{Extrapolation and learning can go together: Studies in cognitive science, psychology, and neuroscience}

\vspace{-.2cm}

% \ry{another option is to move some of these items to an appendix and refer the reader there -- i would imagine the lengthy ones that are less relevant to our main argument -- e.g., maybe we can move function learning?}

Extrapolation is studied widely in relation to different learning tasks, in cognitive science, psychology, and neuroscience. Moreover, many aspects of human visual cognition and visual behavior involve extrapolation. Here, we briefly review that literature to provide a better understanding of relevant extrapolation frameworks and their relation to learning, as such frameworks have been absent from mathematical studies on generalization of image classification models. At the same time, we note that our findings may contribute not only to the mathematical understanding of deep learning, but also to the psychology and cognitive science literature as image classification is not yet considered an extrapolation task. In recent years, there has been an intimate interaction between the fields of psychology and machine learning where findings in one field often prompt new studies in the other field \citep{peterson2021using}. We consider implications of our study in both fields.

% \ry{good?} \jm{I think it's good. Could edit it to be shorter if needed but I like the overall message} \ry{great! we might already be within the 4 page limit, but definitely a good idea to make it brief if we are over the 4 page}

% \ry{Image classification is not yet considered an extrapolation task in the cognitive science literature. Can our study contribute to the cognitive science literature, too? Should we mention that possibility?}

% \ry{following is in progress - it's chaotic right now -- we can reconsider order of items}

% \cite{lotter2020neural} says humans are far better than DNNs in few-shot learning and in extrapolation across domains.

% check if DiCarlo has any papers on extrapolation?

% \ry{"many aspects of human visual cognition and behavior involve extrapolation" -- if this statement (or some version of it) sounds correct, we can mention this in the abstract, introduction, and conclusion, as well}

{\bf Convex hulls.} \cite{webb2020learning} study the extrapolation capabilities of neural networks with the motivation "that human reasoning involves the ability to extrapolate." They consider the distance to the convex hull of training set and conclude, from experimental results, that ``the ability of neural networks to extrapolate tends to degrade as a function of" that distance. To facilitate extrapolation studies, they provide a new dataset called Visual Analogy Extrapolation Challenge (VAEC) that is constructed from objects, such as a rectangle, that vary in brightness, size, and location along the horizontal and vertical axes. They use an autoencoder (with 3 convolutional layers) to generate a low-dimensional embedding for images, then train a single-layer LSTM network to extrapolate from a sequence of embeddings (e.g., for images depicting rectangles of different size), but they do not aim to classify contents of images. Moreover, they do not consider the task of standard image classification models as extrapolation. Instead, they assume that "generalization exhibited by contemporary neural network algorithms is largely limited to interpolation between data points in their training corpora." As we reviewed in previous section, testing samples of standard image classification datasets are considerably outside the convex hull of training sets in the pixel space and in the feature space learned by networks, and therefore, that assumption is not accurate. In fact, classification of standard image datasets such as CIFAR-10 requires extrapolation similar to the proposed VAEC dataset. 

{\bf Extrapolation in human visual behavior.} Extrapolation is studied in the visual domain, and multiple studies have shown that humans extrapolate future positions of objects in motion, based on the past trajectory of an object. Intriguingly, extrapolation can lead to visual illusions, such as the flash-lag effect, where a flash that is physically aligned with a moving object appears to lag behind the moving object. It is proposed that this occurs because extrapolation from the object’s current position to a new predicted location causes the flash to be perceived as behind the object in motion \citep{nijhawan1994motion}. Intriguingly, recent research has suggested that extrapolation may occur early on in the visual hierarchy \citep{van2019predictive}. Extrapolation also occurs in the motor system (the part of the nervous system that supports movement), and researchers have found a correlation between eye movement (saccade) latencies and the expected position of a target, such that a longer saccade latency leads to a greater displacement in the expected target position \citep{van2018motion}. Intriguingly, extrapolation may depend on the degree of noise in sensory inputs \citep{khoei2013motion}, and humans estimate and compensate for the degree of uncertainty in a moving source \citep{warren2012visual}. In these studies, often ``time" is the dimension of extrapolation, however, there are studies that focus on geometric aspects of visual extrapolation, e.g., extrapolating to estimate the shape of contours. \citep{singh2005visual,fulvio2014visual}.

% \ry{I love this piece. Some reviewers might not know what motor system is and they might not bother to look it up. Perhaps we can explain it in parenthesis -- perhaps something like the following? "the set of central and peripheral structures in the nervous system that support movement functions."}
% \jm{added a short parenthetical thing about the motor system}

{\bf Extrapolation by animals.} There are studies on extrapolation performed by animals, e.g., ``mice extrapolate the geometric information relative to the boundaries of a maze and use it to navigate also in the presence of a prominent cue" \citep{fellini2011geometric}. \citet{poletaeva2015extrapolation} investigate the distinction between the ability to anticipate reward on the basis of extrapolation, as opposed to instrumental learning.% There are also studies on visual object recognition behavior of primates \citep{rajalingham2018large}, but those studies have not considered the extrapolation aspects of object recognition.%, but object recognition is not yet viewed as an extrapolation task in that context.
%\jm{I am not sure about the Poletaeva reference here, will look at and see if the sentence can be rephrased. Just because we didn't yet define instrumental learning, not sure if comp sci ppl know what it is or not}\ry{Rephrasing sounds good! Comp sci ppl would not exactly know what instrumental learning is, but I just want to tell them that there is some kind of learning based on extrapolation and there's some other kind of learning}
% \jm{Sure. I think we also need to say why the rajalingham2018large ref is relevant, or remove it} \ry{makes sense. I removed it because their work is not about extrapolation. I think as a matter of literature review, ultimately in our main paper (extension of the workshop paper), we should talk about such papers in another section (separate from the extrapolation section).}

{\bf Function learning.} A straightforward application of extrapolation is function learning. Once one learns a function, the function can be applied to any input within the function's domain (inside and outside the \hull of data used for deriving the function). Function learning can be used to generalize from previous experience in cases where not all possible states can be observed \citep{wu2018generalization}. 

There are extensive experimental results on how humans extrapolate, e.g., providing a set of inputs and outputs of a function to a human, and then asking them to guess the output for some unseen inputs outside \hull \citep{delosh1997extrapolation}. In computational cognitive psychology, there is a literature on building mathematical models that can emulate human interpolation and extrapolation behavior \citep{lucas2015rational,bott2004nonmonotonic,villagra2018data,kalish2013learning}. Notably, \cite{griffiths2008modeling} used Gaussian processes and basis functions to model human function learning. Computational models have also been extended to explain how neural representations, particularly in the perceptual and motor systems, solve extrapolation problems in function learning \citep{guigon2002neural}.

{\bf Graphics and visual patterns.} Extrapolation is studied in the context of graphics, specifically graphics and visual patterns derived from hand drawings \citep{ellis2018learning}.% \cite{storrs2021learning} studied spatial extrapolation in image space to complete partial images.

{\bf Category learning.} Extrapolation is studied in the context of category learning (i.e., classification) \citep{mcdaniel2005conceptual,navarro2012anticipating,schlegelmilch2020cognitive}. Notably, \citet{sewell2011restructuring} consider a bounded 2D domain and 2 classes of scattered points within a portion of that domain , i.e., \hull. They then study how decision boundaries that separate the points extend throughout the domain (depicted in their Figures 2 and 3) from the perspective of different cognitive architectures. This setting closely relates to how we study our deep learning functions. Intriguingly, \cite{silliman2021extrapolation} considers contrastive category learning  and exemplars with the idea that humans ``develop caricaturized representations (i.e., ideals or extreme points) to support successful discriminative classification". This idea connects with our next point about convex hulls because creating extreme points in the representation space entails enlarging the convex hull of observed data. 
% \jm{I am not understanding the sewell2011restructuring ref, I don't think they have any neural data. reading it, maybe we can rephrase the part saying that they consider it from the perspective of the brain. I rephrased it as "cognitive architectures". It could also be different "computational models" if cognitive architectures is too specialized}\ry{yes, what you have is more accurate and definitely better.}

In an exemplar model of categorization, new inputs are compared with specific instances saved in one's memory, but human visual behavior may be more complicated than that \citep{poggio2004generalization}. There is a long standing literature suggesting that humans learn to categorize objects, situations, etc using prototypes \citep{rosch1978cognition,smith2002distinguishing} where a prototype is not necessarily a real instance (i.e., exemplar) but can be some idealized or average representation. Since deep learning models are not interpretable, it has been suggested to incorporate a prototype model in their structure \citep{chen2018looks}, but, so far, methods to derive the prototypes require a model that is already trained. In other words, deriving \underline{useful} prototypes is a challenging task to begin with, while deriving \underline{invariant} prototypes may be even more difficult. In cognitive psychology, it is known that when using a prototype model, we should not expect invariance with respect to the prototypes because the inputs we encounter can have a large variety and there usually will be a competition among several plausible prototypes \citep{appiah2008experiments}. For example, in object recognition, images of cars can have a large variety with respect to the shape of the car, colors, shades, views, etc, some of which might look like other objects such as trucks, etc. And any new car image that we encounter may have some novelty that is not captured in previous car images used to build our prototype. Therefore, we may have to extrapolate from our previously built prototypes in order to classify new images.
% \jm{I rephrased this part "In other words, deriving useful prototypes is a challenging task to begin with, let alone, deriving invariant prototypes", check if looks ok}\ry{this is definitely better!}

% \ry{please see if this makes sense. please feel free to modify it. we can also drop it all together if you think it can be considered controversial or distract the reviewer from the main message.}
% \jm{if mentioning prototypes, I feel like exemplars should at least briefly be mentioned too. but not sure if space}
% \jm{I don't really get the point about the chen paper, maybe unpack more} \ry{Please see if the above makes better sense.}
% \jm{seems fine. I think the "we should not expect invariance" part is a little confusing still, maybe unpack it more. Used to thinking of invariance as a representation of the same visual item in different angles, colors, etc. not sure if that's what you mean or if the prototypes are invariant} \ry{helpful comment :) I tried to unpack more and explain better. See if it makes better sense now}\jm{edit looks good!}

% Later, we will explain how our extrapolation framework and the distance to the convex hull can put this issue in perspective.

{\bf Confidence in decision making.} Extrapolation is studied in relation to confidence in predictions made by a model. Indeed, there are experimental studies that report humans are less confident about their predictions when they extrapolate rather than interpolate, and moreover, there is a correlation between human confidence and the correctness of predictions \citep{stojic2018you}. %Extending the same notion to extrapolation, the farther we go outside the \hull (especially in the feature space), one could be less confident in the correctness of classifications of a model. 
This relates to the failure of deep networks in out-of-distribution detection, defying the notion of learning when they mis-classify out-of-distribution images with high confidence (e.g., an object recognition model classifying a radiology image of liver as a truck with 100\% confidence) which we will explain later.

{\bf Pure and applied math literature.} There is a rich mathematics literature on extrapolation, which has a wide range of applications. One can learn about extrapolation in introductory numerical analysis books \citep{ascher2011first} as well as in more theoretical fields such as holomorphic approximation and algebraic geometry. In some problems, dimension of extrapolation is time, for example, when predicting the value of stock market for the future. In many other studies, dimension of extrapolation is extending the domain of a function, e.g., in approximation theory \citep{sidi2003practical,fefferman2005interpolation}. There are classic problems that consider how a function derived on a compact set can be extended into a larger domain, e.g., Whitney's Extension Problem \citep{fefferman2006whitney}.% \ry{Added the last sentence here.}

\vspace{-.2cm}

\section{Extrapolation framework can answer open questions about deep learning}

% \ry{we can mention each time very briefly and move the main text to an appendix.}

% \ry{I'll make this change at the end when all pieces are set.} \jm{sounds good, yes, I think this should be moved to appendix}

\vspace{-.2cm}

Adopting the framework, that image classification models are extrapolating within the specific task defined by their training set, has the potential to provide novel answers to some of the open research problems about deep learning. Appendix~\ref{sec:appx_answers} briefly mentions some of these problems which we plan to study in our future work. Also, Appendix~\ref{sec:appx_gen+ext} explains how generalization can be defined in relation to interpolation and extrapolation.

\vspace{-.2cm}

\section{Conclusions}

\vspace{-.2cm}

We proposed a new extrapolation framework where the functional task of image classification models is considered extrapolation (in the pixel space and in the feature space derived from images) within the specific scope defined by the training set. We reviewed the literature in cognitive science and psychology to see how extrapolation is commonly studied in relation to learning, and how many aspects of human visual behavior involve extrapolation. We then explained how our novel extrapolation framework can provide new answers to some of the open research problems about deep learning including about their generalization, over-parameterization, etc.

% We proposed a new extrapolation framework where the functional task of image classification models is considered extrapolation (in the pixel space and in the feature space derived from images) within the specific scope defined by the training set. Studying the geometry of image datasets establish that interpolation framework is not adequate for mathematical understanding of image classification models. While extrapolation can refer to any classification that is outside the \hull (i.e., testing samples related and unrelated to the scope defined by training set), the definition of extrapolation can be narrowed down to the scope of training set. Extrapolation framework has not been adopted so far to study the generalization of image classification models, however, extrapolation and learning can go together as they have a rich literature in cognitive science, psychology, and neuroscience. In that literature, extrapolation can have different meanings.

% \ry{this might need to go to introduction - or it might need to be changed/removed.}

% However, we can narrow down the definition of extrapolation to the specific scope defined by the training set. This approach of extrapolating within a specific scope is commonly used in cognitive science and psychology, as we reviewed in this paper. We then explained how this extrapolation framework can provide novel answers to open research problems in deep learning, questions that have remained unanswered so far within the interpolation framework.

\vspace{-.2cm}

\ack{The authors thank anonymous reviewers for helpful comments. R.Y. thanks Dianne O'Leary for helpful comments on appendices. R.Y. was supported by a fellowship from the Department of Veteran Affairs. J.M. was supported by the National Institutes of Health. The views expressed in this manuscript are those of the authors and do not necessarily reflect the position or policy of the Department of Veterans Affairs or the United States government.}

%\jm{I added my funding, do you think I need to add the specific grant number that Hedy Kober supports me on?}
%\ry{I don't think you need to provide a grant number since you said you have worked on this on your own time, i.e., weekends, etc. And most likely, Kober's grant is not related to the topic of this paper. But, its totally fine to acknowledge NIH.}
%\jm{Ok sounds great! Yep exactly}

\bibliographystyle{plainnat}
\bibliography{refs,refs-cognitive}

\clearpage

\appendix

\section{Projections of testing samples to \hull} \label{sec:appx_projections}

Here, we provide several more examples of testing samples of CIFAR-10 dataset and their projections to the \hull in the pixel space.
% \jm{"their directions" may not be grammatical, not sure how to rephrase it. their transformations to? also, below may say "these perturbations}

\vspace{-.2cm}

\begin{figure}[h]
\begin{center}
    \begin{minipage}[b]{.85\columnwidth}
    \centering
    (original) \raisebox{-.5\height}{\includegraphics[width=.20\columnwidth]{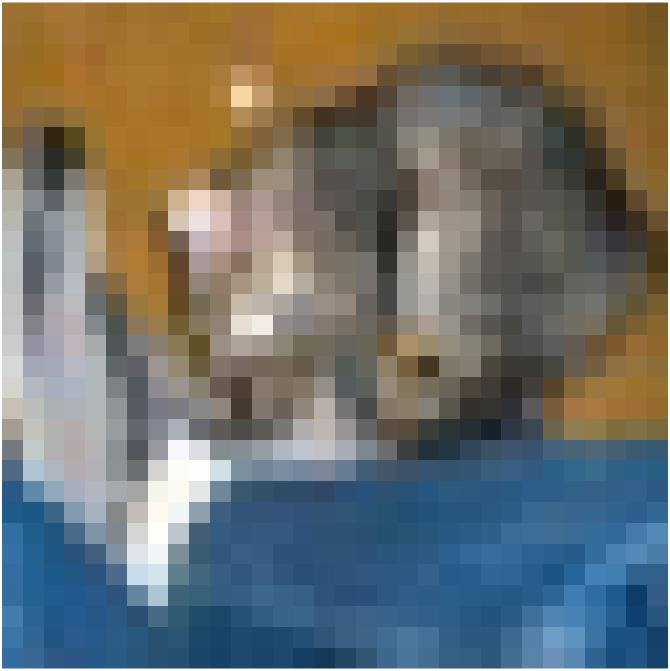}} {\vspace{.2cm} -}
    \raisebox{-.5\height}	{\includegraphics[width=.20\columnwidth]{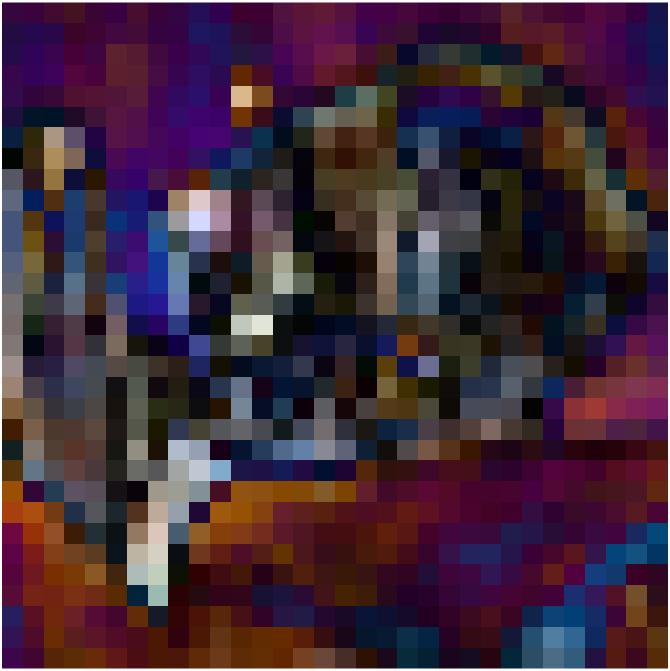}} =
    \raisebox{-.5\height}	{\includegraphics[width=.20\columnwidth]{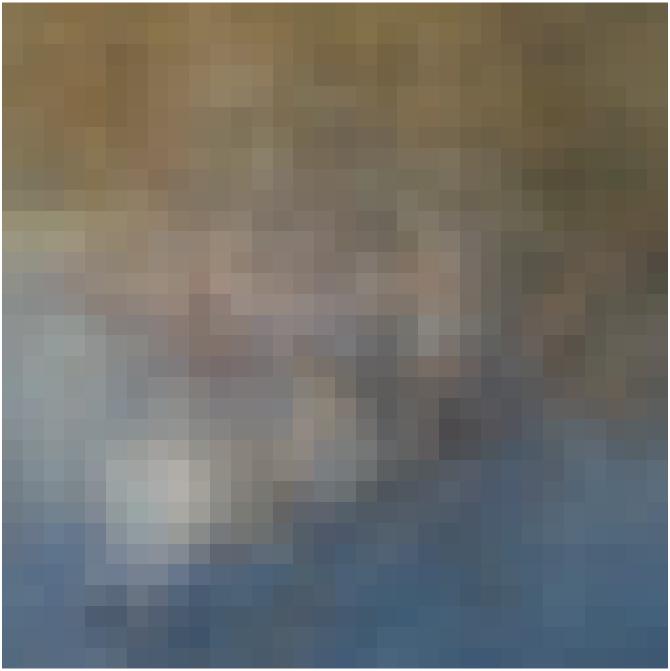}} (on \hull)
    \end{minipage}
    \begin{minipage}[b]{.85\columnwidth}
    \centering
    (original) \raisebox{-.5\height}{\includegraphics[width=.20\columnwidth]{figures/cif10_te2orig.jpg}} {\vspace{.2cm} -}
    \raisebox{-.5\height}	{\includegraphics[width=.20\columnwidth]{figures/cif10_te2diff_all.jpg}} =
    \raisebox{-.5\height}	{\includegraphics[width=.20\columnwidth]{figures/cif10_te2oncvx_all.jpg}} (on \hull)
    \end{minipage}
    \begin{minipage}[b]{.85\columnwidth}
    \centering
    (original) \raisebox{-.5\height}{\includegraphics[width=.20\columnwidth]{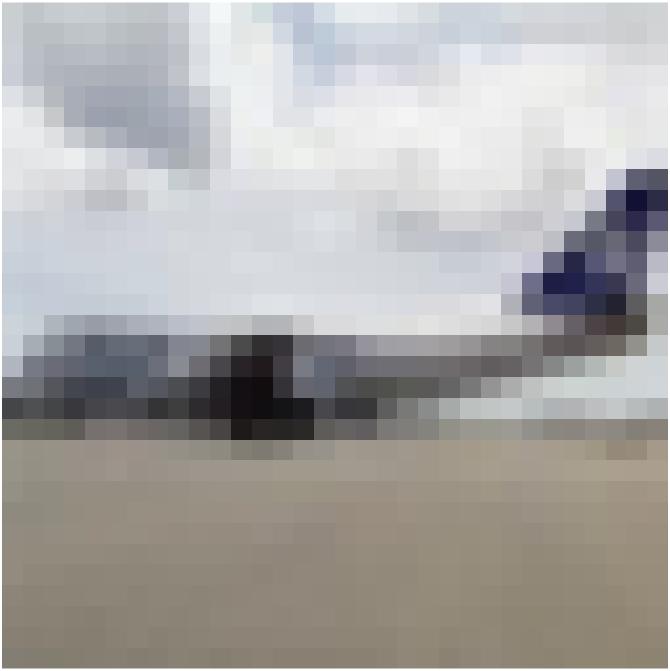}} {\vspace{.2cm} -}
    \raisebox{-.5\height}	{\includegraphics[width=.20\columnwidth]{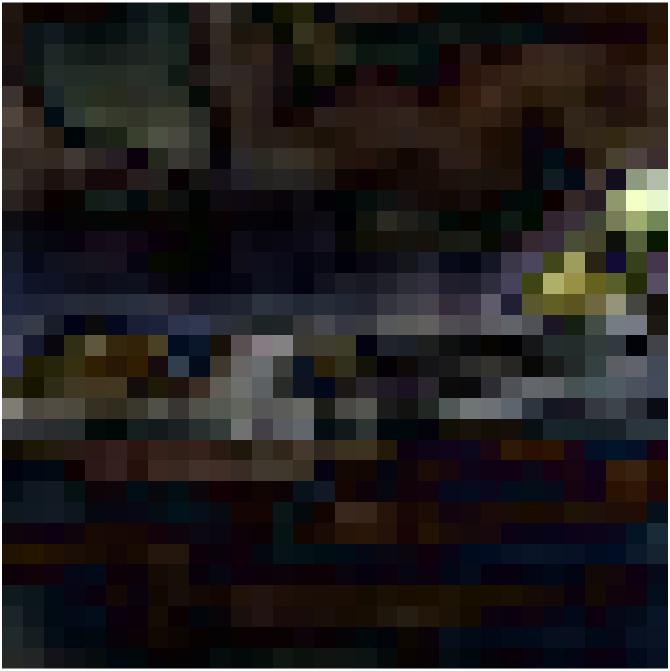}} =
    \raisebox{-.5\height}	{\includegraphics[width=.20\columnwidth]{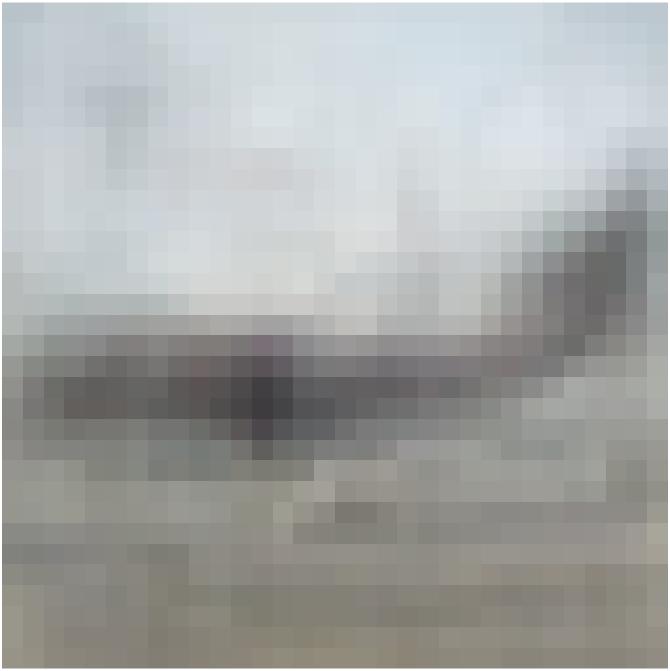}} (on \hull)
    \end{minipage}
    \begin{minipage}[b]{.85\columnwidth}
    \centering
    (original) \raisebox{-.5\height}{\includegraphics[width=.20\columnwidth]{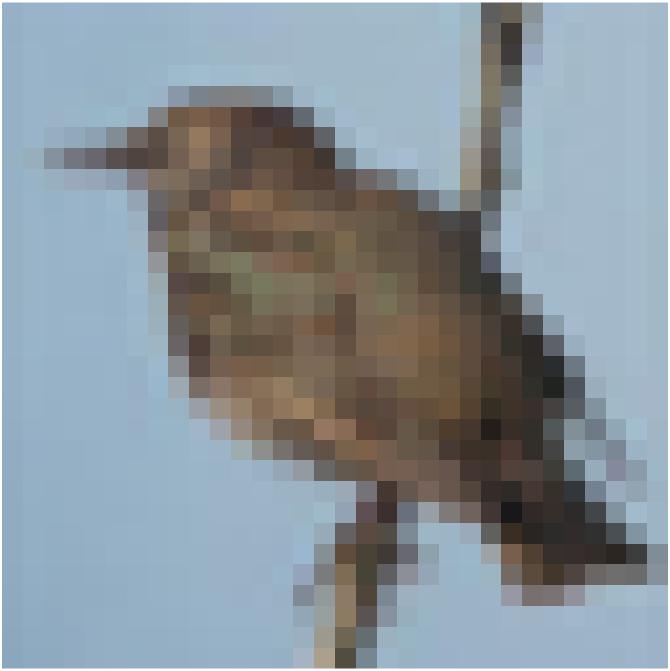}} {\vspace{.2cm} -}
    \raisebox{-.5\height}	{\includegraphics[width=.20\columnwidth]{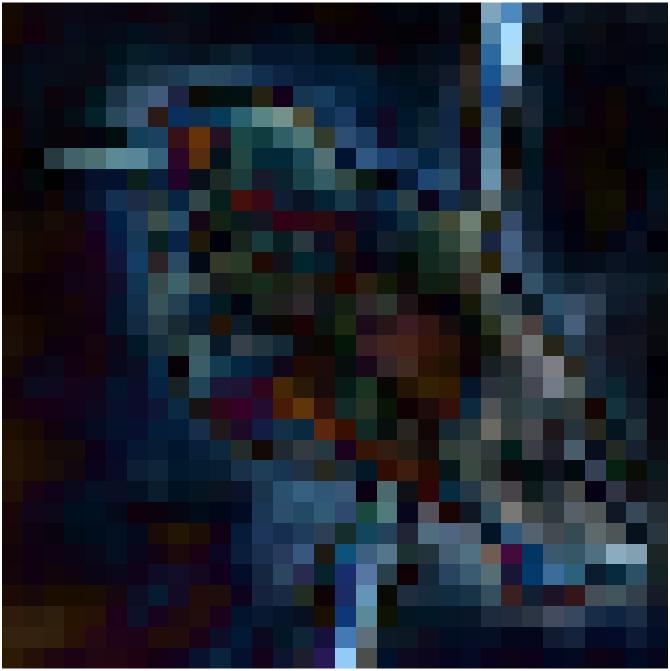}} =
    \raisebox{-.5\height}	{\includegraphics[width=.20\columnwidth]{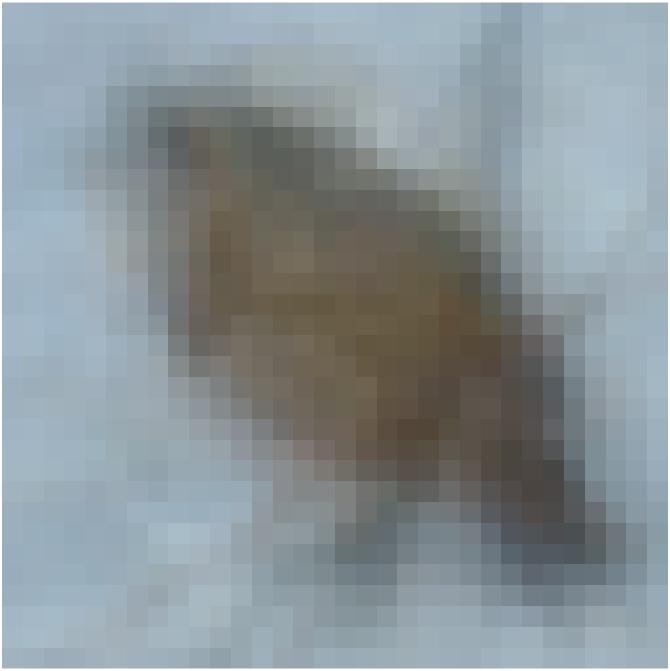}} (on \hull)
    \end{minipage}
    \begin{minipage}[b]{.85\columnwidth}
    \centering
    (original) \raisebox{-.5\height}{\includegraphics[width=.20\columnwidth]{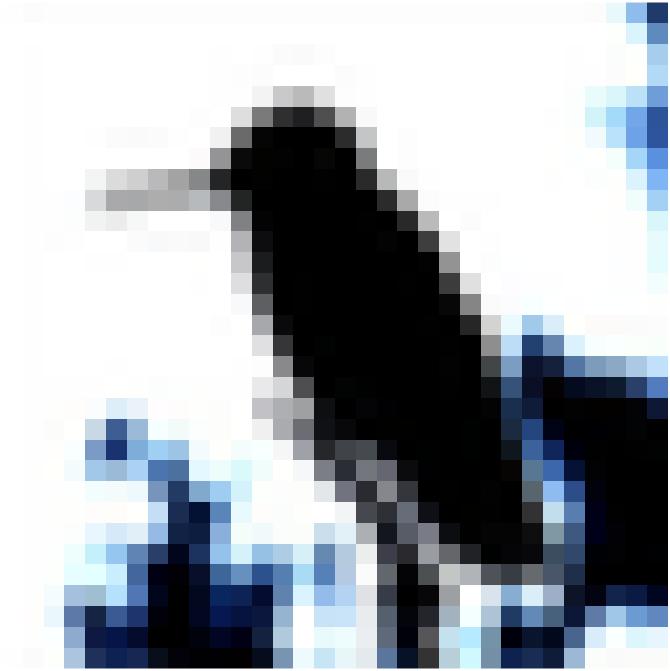}} {\vspace{.2cm} -}
    \raisebox{-.5\height}	{\includegraphics[width=.20\columnwidth]{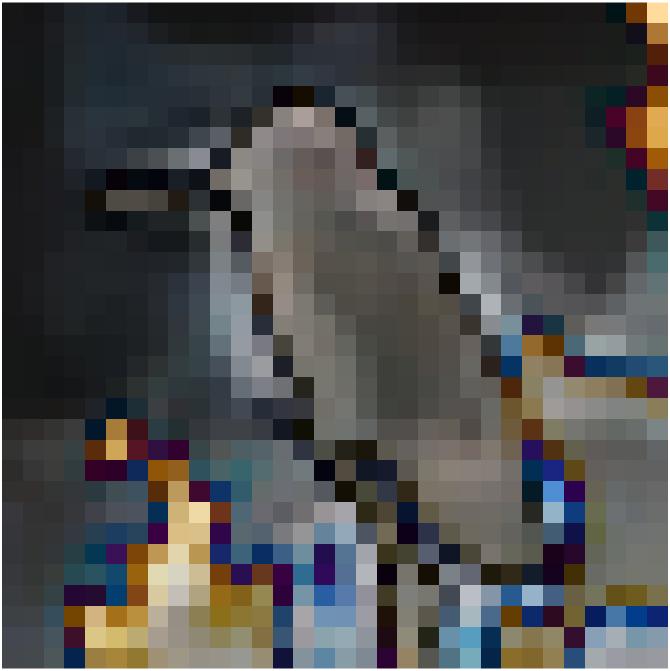}} =
    \raisebox{-.5\height}	{\includegraphics[width=.20\columnwidth]{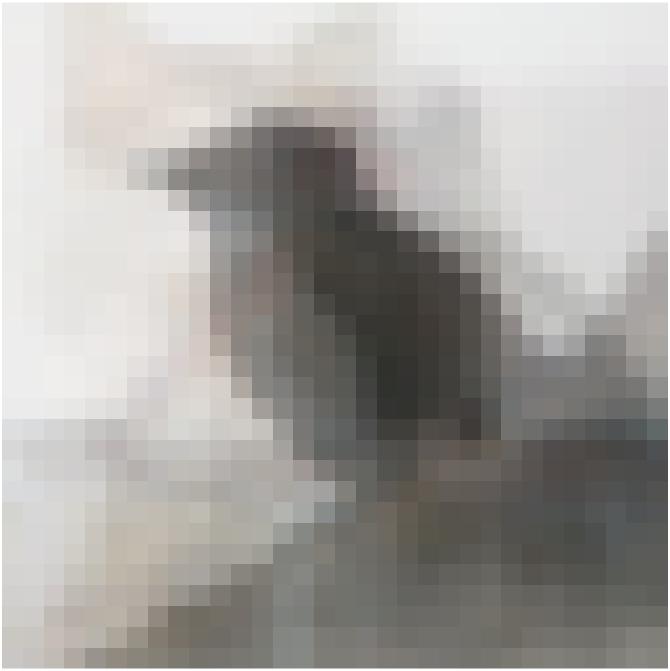}} (on \hull)
    \end{minipage}
    \begin{minipage}[b]{.85\columnwidth}
    \centering
    (original) \raisebox{-.5\height}{\includegraphics[width=.20\columnwidth]{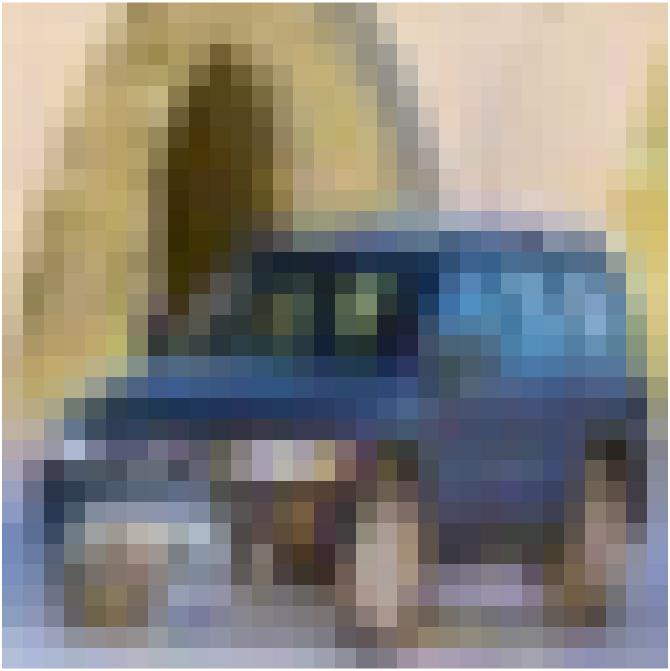}} {\vspace{.2cm} -}
    \raisebox{-.5\height}	{\includegraphics[width=.20\columnwidth]{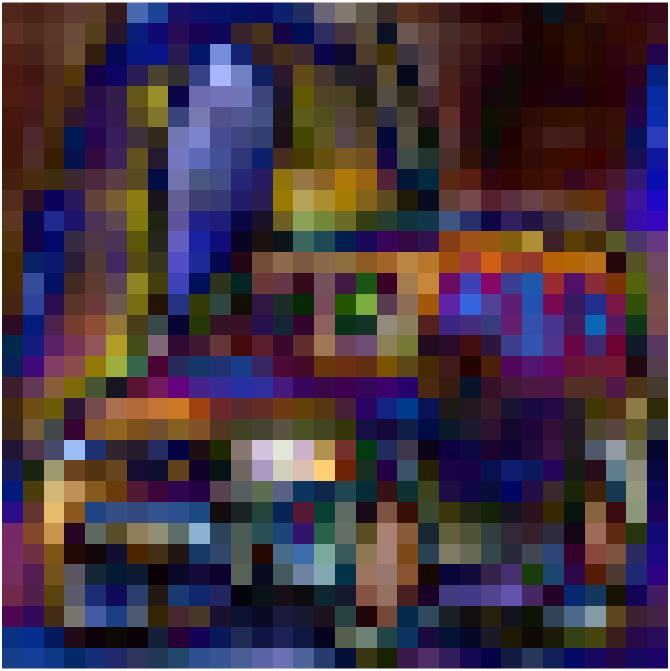}} =
    \raisebox{-.5\height}	{\includegraphics[width=.20\columnwidth]{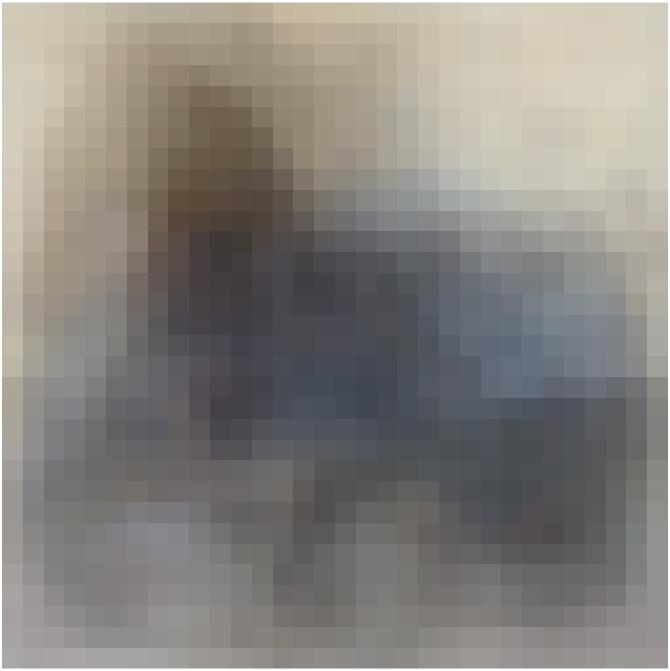}} (on \hull)
    \end{minipage}
    \begin{minipage}[b]{.85\columnwidth}
    \centering
    (original) \raisebox{-.5\height}{\includegraphics[width=.20\columnwidth]{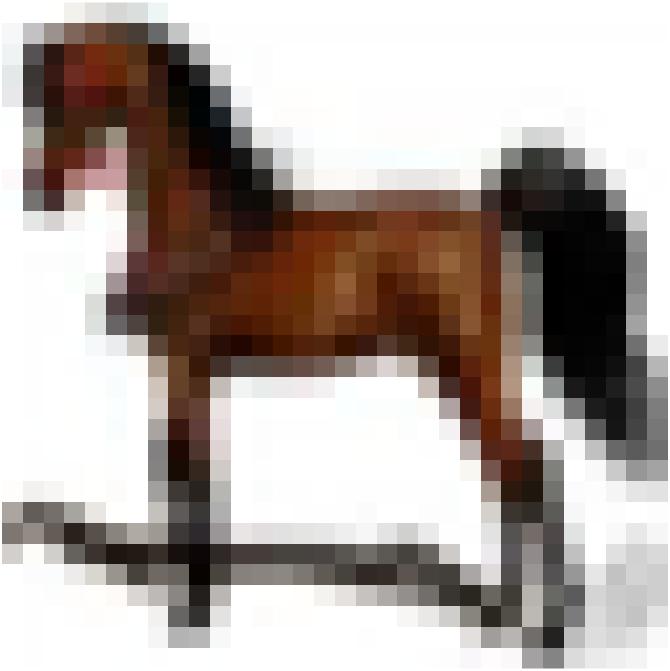}} {\vspace{.2cm} -}
    \raisebox{-.5\height}	{\includegraphics[width=.20\columnwidth]{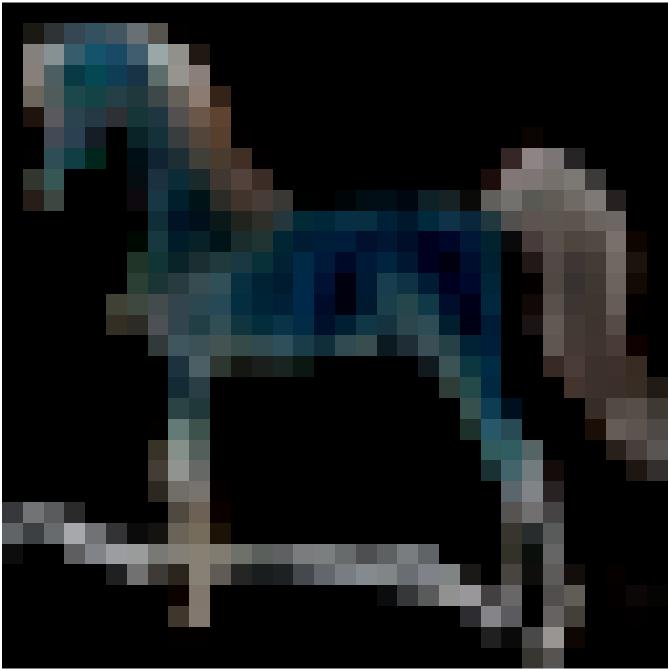}} =
    \raisebox{-.5\height}	{\includegraphics[width=.20\columnwidth]{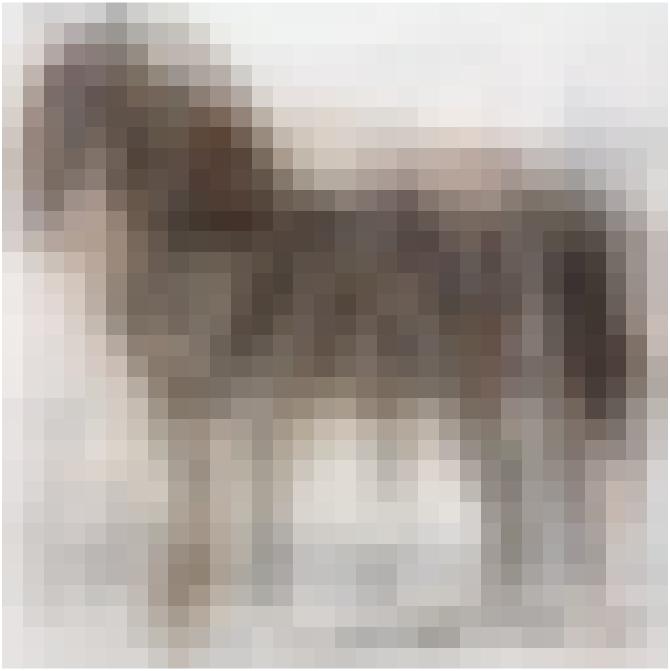}} (on \hull)
    \end{minipage}
\end{center}
\vspace{-.2cm}
    \caption{Minimum perturbation that can bring a testing image to \hull. (left image) original testing image from CIFAR-10, (middle image) what should be changed in the original image, (right image) the projection on the \hull. For each testing sample, the projection from the sample to the closest point on the surface of the hull relate to features and objects of interest in the testing image.}
  \label{fig:direction_to_hull_appx}
\end{figure}

% \section{Extrapolation as the functional task of deep networks} \label{sec:appx_geometry}

% Most interestingly, we see that even in the last hidden layer of a pre-trained CNN trained on CIFAR-10, all testing samples are considerably outside the \hull of that feature space. This implies that the features that a model extracts and learns from images do not transform the functional task of the model into interpolation. In other words, any testing image (that the model encounters) has some novelty that puts the image outside the \hull. For feature selection with wavelets and shearlets, we convolve the images with a wavelet basis such as Daubechies wavelets \citep{daubechies1992ten} or with a shearlet system \citep{andrade2020shearlets}. We combine the resulting coefficients in a matrix and then perform feature selection. For feature selection, we use rank-revealing QR factorization \citep{chan1987rank} which selects the features based on their rank contribution, and also use Laplacian score \citep{he2005laplacian} which selects features based on a nearest neighbor similarity graph. We arrive at similar distance distributions with respect to the resulting \hull's. %The number of features we select can be considered low-dimensional, e.g., for MNIST, all the trends remain similar when we select 50, 64, 100, 200, and 400 features from images (while the pixel space has 784 dimensions).

% \section{Generalization and its relationship with extrapolation} \label{sec:appx_gen+ext}

\section{Possible confusions about definition of interpolation} \label{sec:appx_interextra}

As mentioned earlier, in deep learning literature, many studies solely use interpolation to explain the functional task of deep learning models. Recently two papers summarized those studies \citep{belkin2021fit,dar2021farewell}, emphasizing that deep learning models interpolate.

A confusion may arise that interpolation may be defined as perfectly fitting a set of points. However, whether a model fits a set of points or not, the training set which the model is trained on has a convex hull and any classification a model does outside that convex hull will satisfy the definition of extrapolation, and classifications outside the convex hull of training set cannot be explained by mere interpolation.

{\bf An imaginary feature space:} It may be argued that there is a feature space where testing samples are contained in the convex hull of training set, and in such space, functional task of the model is interpolation. However, there is no evidence for existence of such space as we have investigated empirically \citep{yousefzadeh2021hull}. All evidence suggest that testing samples are outside the convex hull of training sets even in feature spaces that trained models extract from images.

{\bf Generalization and functional task of models:} Fitting the training data does not imply generalization. Rather, models are useful only if they can classify unseen images with some acceptable accuracy. Merely fitting a model to a training set does not lead to good generalization. In fact, image classification models have the capacity to achieve near zero training loss, even when training samples are labeled randomly and even when contents of images are replaced with random noise \cite{zhang2016understanding,zhang2021understanding}, both cases leading to testing accuracy similar to random guess. A model may fit a correctly labeled training set, yet generalize worse than random guess.

When we study the functional task of image classification models, we are focused on generalization of the models and their classification of unseen data, i.e., testing samples. When testing samples are outside the convex hull of training set, in the pixel space or in the feature space, a trained model would need to extrapolate outside the convex hull in order to classify them.

{\bf High-dimensionality of data:} It may be argued that high-dimensionality of data leads to testing samples falling outside the convex hull of training set, the extent of extrapolation is negligible, and therefore, their functional task may be considered interpolation. This is not supported by empirical results, in pixel space and also in feature space. The last hidden layer of the trained model we described earlier has only 64 dimensions while CIFAR-10 has 50,000 training samples. Even in that feature space, distance of testing samples to the hull is not negligible. A testing image has to significantly change in order to reach the convex hull of training set, and those changes relate to the objects of interest in images.

\begin{figure}[h]
  \centering
  \includegraphics[width=0.25\linewidth]{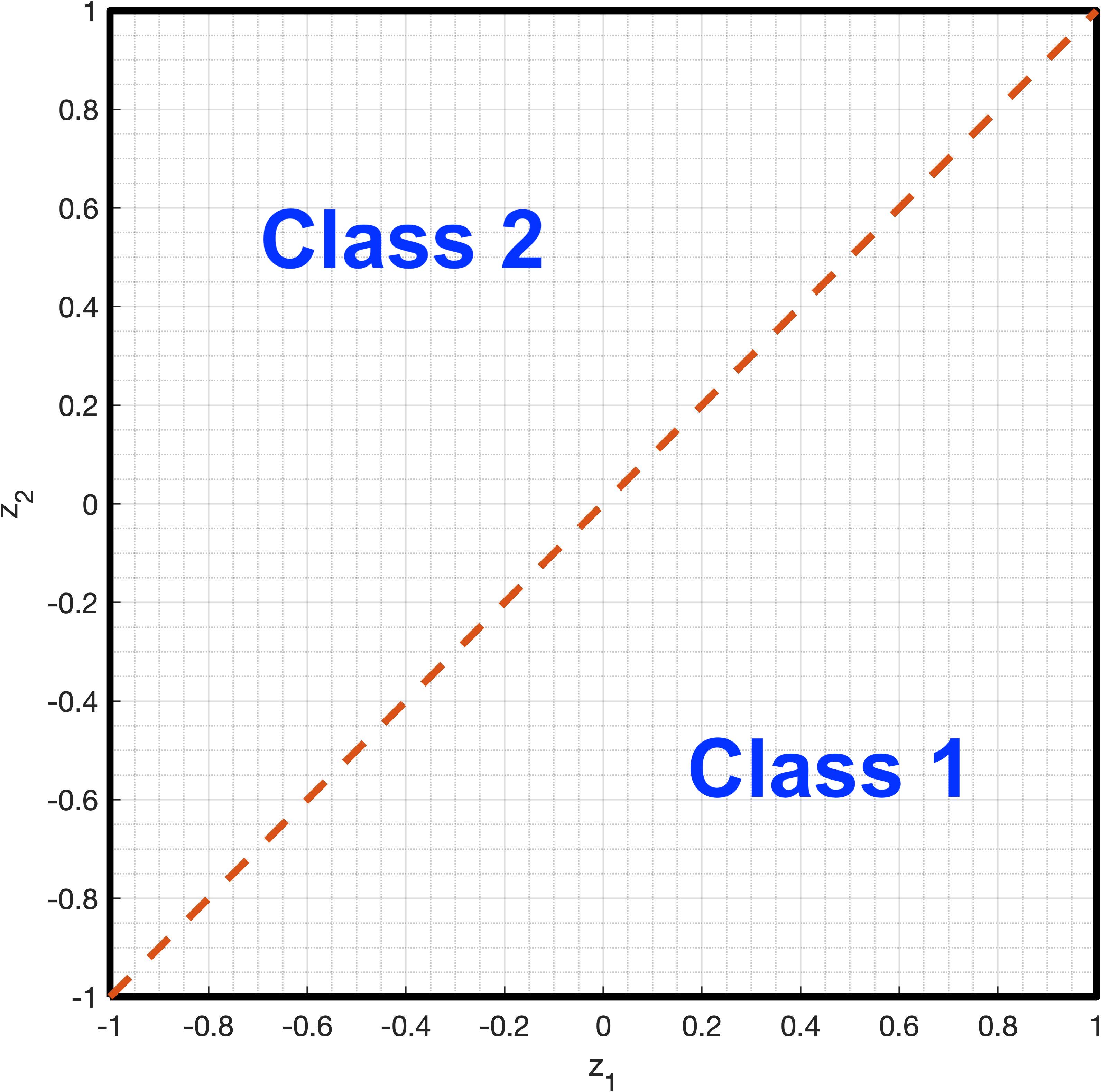}
  \caption{Partitioning of the output layer of a classification model is predefined. Max operation and sort operation on a vector does not imply interpolation.}
  \label{fig:output_partition}
\end{figure}

{\bf The output layer:} The layer after the last fully connected layer is the output of the model whether we apply softmax to it or not. The class(es) with the largest output value will be the classification of a model for a particular input. That does not imply interpolation as a functional task. Decision boundaries of a model are predefined for the output of the model. Consider, for example, a model with binary output, $\boldsymbol{z}$. Decision boundary in the output layer is defined by the line $z_1 = z_2$, as shown by the red line in Figure~\ref{fig:output_partition}. In the space of $\boldsymbol{z}$, the region $z_1>z_2$ is the partition for class 1, and the region $z_1<z_2$ is the partition for class 2. The output layer of a model is like a dart board with predefined partitions/regions. When a dart lands on a specific region, that region defines the result. The computation performed on the output of the model (whether softmax is applied to it or not) is either a max operation or a sort operation, none of which imply interpolation.

{\bf Regression:} In regression, when fitting a function to a set of points, a loss function is used that measures the discrepancies between the points and the function values. Least squares problems are one type of defining such function in which all discrepancies are squared and summed \citep{golub2012matrix}. When there is no discrepancy between the points and the function values, the loss function will be minimized to zero, and the regression function would pass through all the points and interpolate them.

When a function is obtained that interpolates/fits a specific set of data points (or even when the function does not perfectly fit the points), the function may be used both for interpolation purposes and for extrapolation purposes. Using the function within the convex hull of training set would be considered interpolation, and using the function outside that convex hull would be considered extrapolation. When it comes to the generalization of a function, the distinction between interpolation and extrapolation is defined with respect to the convex hull of data used to derive the function.

% Similarly, for classification functions, the distinction between interpolation and extrapolation is defined in relation to the convex hull of data used to derive the function, i.e., the training set. 

{\bf Classification functions:} In deriving a classification function from a set of labeled data points, the goal is to obtain a function that partitions the domain such that data points for each class are separated from data points of other classes. Partitions in the domain are defined by decision boundaries. The classification function will ideally maintain a distance (margin) between the decision boundaries and the data points. A classic example for this is Support Vector Machines (SVMs) in which the goal is to partition the domain such that the minimum margin between the decision boundary and data points is maximized.

% svm figure

% a support vector machine partitioning two sets of points in the domain while maximizing its margin.

For image classification models, it is customary to use the cross-entropy as the loss function when training the models. Cross-entropy, however, does not consider the margin, rather, loss is incurred whenever there is a discrepancy between the output of the model for a sample and the true label for that sample. As long as the output of the model and the true label coincide, no loss will be incurred. \cite{elsayed2018large} considered adding an extra penalty to the cross-entropy to push the decision boundaries away from the samples and showed that it can lead to better generalization of the models. \cite{zhang2018mixup} also showed that one can average pairs or triplets of training samples and their labels, and train the models on those mixed images. This is defining the location of decision boundaries in the pixel space.

In this setting, minimizing the cross-entropy loss does not necessarily imply interpolation. It merely implies that our model has been able to partition the domain such that training samples of each class are perfectly separated from other classes. When the model receives a testing sample, the sample would map to one of the partitions in the domain (or on a decision boundary between the partitions), and classification of the model is based on that mapping. 
Regardless of the shape of the partitions in the domain, any classification that the model does outside the convex hull of training set, from the mathematical perspective, can be considered extrapolation.

% The domain of the model might be convex (e.g., a hyper-cube), but partitions within the domain are not necessarily convex. Although a perfect partitioning may minimize the cross-entropy loss to zero, this does not necessarily lead to interpolation.

\section{Using the extrapolation framework to provide new answers to deep learning mysteries} \label{sec:appx_answers}

\subsection{Over-parameterization of image classification models}

Deep learning models are highly over-parameterized and there is not a clear justification for this over-parameterization. Notably, there are infinite number of minimizers that make their training loss close enough to zero, while many of those minimizers lead to models that generalize very poorly. There is not an independent way to verify which minimizers lead to good generalization, unless we test the resulting model on some validation/testing set. From another perspective, extensive over-parameterization of these models provides the capacity for them to reach zero training loss even when training images are randomly labeled, or when the content of images are replaced with random noise \citep{zhang2016understanding}. So far, there is not a clear explanation in the literature about these questions as reiterated recently by \citet{zhang2021understanding}. However, once we adopt the extrapolation framework, we can see that all these phenomena can be explained within that framework. One can show for general decision boundaries that over-parameterization is a necessary condition for having control over the extensions of decision boundaries outside the \hull. Moreover, a model that is sufficiently over-parameterized\footnote{Over-parameterization of a model is sometimes evaluated by comparing the number of training samples with the number of parameters in a model. But, that comparison does not consider the distribution of training samples, e.g., one can inflate the number of training samples by adding samples that are almost redundant and not useful for generalization. Here, we use a more clear definition for over-parameterization. We consider a model to be over-parameterized when it has more parameters than necessary to achieve near zero training loss, i.e., more than the minimum number of parameters required for the model to correctly classify all its training samples.} will have the capacity to partition the domain in infinite number of ways. Partitioning the domain in a specific way would require using a specific training regime, answering another mystery about deep learning. Once we adopt the extrapolation framework, we can draw from the centuries-old applied math literature (on approximation theory, domain partitioning, and classification) instead of creating new theories such as the theory of double descent that tries to understand deep learning via interpolation.% and have not been able to solve the mysteries of deep learning \citep{mystery2020arora}.

\subsection{Relating the extent of extrapolation to the advantage of deep learning}

There are certain learning tasks where DNNs have not shown any clear advantage over simple models \citep{rudin2021interpretable}, e.g., for the FICO explainability challenge at NeurIPS 2018 \citep{fico2018challenge}, the best accuracy was obtained by a simple model \citep{chen2018interpretable}. When we investigate the geometry of these datasets, we see that the extent of extrapolation is negligible for such learning tasks, and also considerable portion of their testing samples (about 50\% for FICO dataset) are within their \hull. We note that for image datasets such as CIFAR-10, the extrapolation is significant even in the 64-dimensional space of last hidden layer.

{\bf CIFAR-10 dataset:} All testing samples of this dataset are outside the convex hull of its training set. The diameter of the convex hull of the training set is 13,621, measured in Euclidean distance in the pixel space. Figure~\ref{fig:dist_cvx_cif10} shows the distribution of distance to convex hull for all testing samples of this dataset. Distance of testing samples to convex hull of training set ranges from 1\% to 27\% of the diameter of convex hull (with average ratio of 10\%).

\begin{figure}[h]
  \centering
  \includegraphics[width=0.45\linewidth]{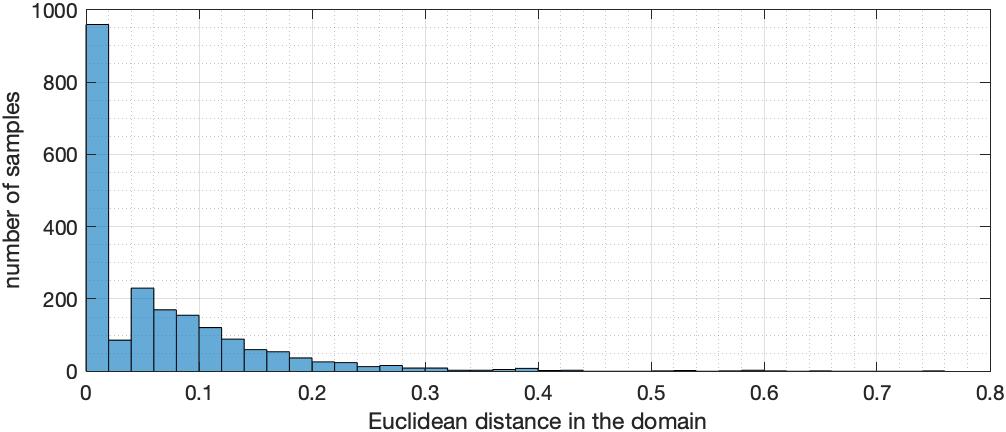}
  \caption{Variations of distance to convex hull of training set for all testing samples of FICO dataset.}
  \label{fig:dist_cvx_fico}
\end{figure}

{\bf FICO dataset:} We randomly pick 20\% of this dataset as testing set and use the remaining 80\% as the training set. We report the Euclidean distances in the domain. The diameter of the convex hull of training set is 4.15. Of all the testing samples, only 54\% of them fall outside the convex hull of training set. Figure~\ref{fig:dist_cvx_fico} shows the distribution of distance to convex hull for all testing samples. The distance of samples that fall outside the convex hull, ranges between 0\% to 18\% of the diameter of the convex hull (with average ratio of 1.5\%).

\begin{figure}[h]
  \centering
  \includegraphics[width=0.45\linewidth]{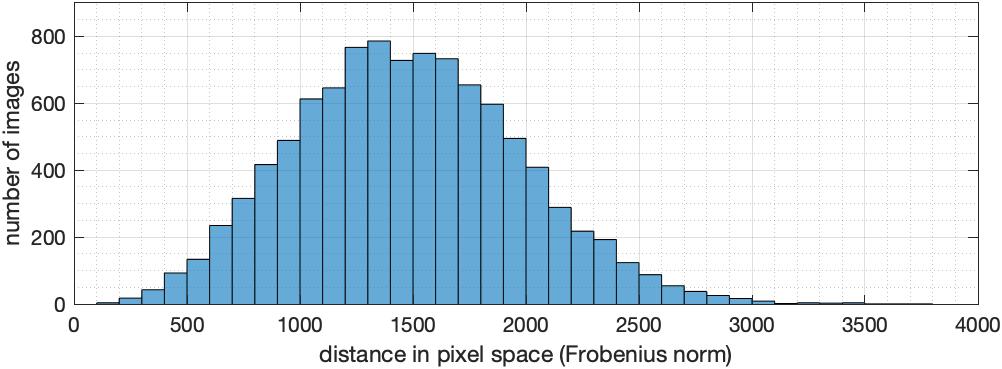}
  \caption{Variations of distance to convex hull of training set for all testing samples of CIFAR-10 dataset.}
  \label{fig:dist_cvx_cif10}
\end{figure}

{\bf Comparison:} It is clear that extent of extrapolation is much less for the FICO dataset compared to CIFAR-10. Almost half of testing samples of FICO dataset are inside the convex hull and no extrapolation is required to classify them. Most testing samples of FICO that are outside the convex hull are relatively close to it. On the other hand, testing samples of CIFAR-10 are entirely outside the convex hull of its training set (even in the feature space). On average, the extent of extrapolation required to classify testing samples of CIFAR-10 is about 6 times larger then the extent of extrapolation required for the FICO dataset. Overall, we can say that the extrapolation required for the FICO dataset is negligible in comparison to CIFAR-10.

% We mentioned earlier that empirically, deep networks do not have any advantage for learning the FICO dataset, while they are the best model for learning the CIFAR-10 dataset.

The above contrast between the 2 datasets matches the contrast we mentioned earlier about the advantage of deep networks in learning these datasets. When the extent of extrapolation is negligible (1.5\% for the FICO dataset), deep networks do not have an advantage over simple models as reported in the literature. However, when the extent of extrapolation is significant (10\% for the CIFAR-10 dataset), deep networks have a clear advantage over simple models.

\subsection{Out-of-distribution detection}

\begin{figure}[h]
  \centering
  \includegraphics[width=0.52\linewidth]{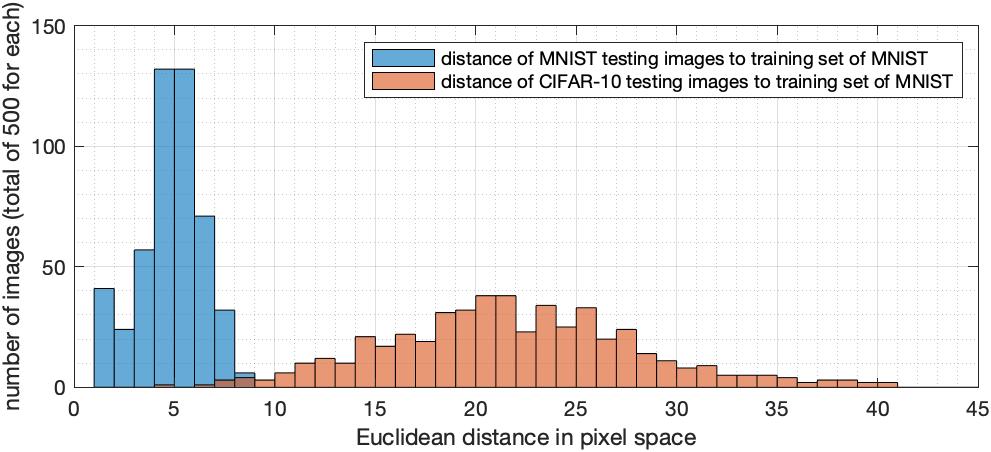}
  \caption{Variations of distance to convex hull of training set in the pixel space for testing samples of CIFAR-10 and MNIST datasets. Distance to \hull of MNIST can separate most testing samples of these two datasets.}
  \label{fig:ood_cif10_mnist}
\end{figure}

One of the shortcomings of deep learning models stems from the fact that they can make classifications with high confidence about images completely unrelated to the scope of their training set. For example, a model trained for object detection tasks such as the one defined by CIFAR-10 dataset can receive a radiology image and classify it with 100\% confidence as a truck or some other object. This is due to the fact that decision boundaries of a model extend throughout the domain outside the convex hull of training set, and as long as a testing sample is not too close to the decision boundaries, the model can be misguidedly confident about its classification even if it is too far outside the convex hull of training set. Our empirical analysis shows that there is very little overlap between the convex hulls of unrelated image classification datasets, e.g. MNIST and CIFAR-10 datasets as shown in Figure~\ref{fig:ood_cif10_mnist} which is a common case study in out-of-distribution detection studies. Such overlap is relatively small in the pixel and even smaller in the feature space. Therefore, the distance to \hull may be used as a straightforward measure for out-of-distribution detection.

\section{What we talk about when we talk about generalization} \label{sec:appx_gen+ext}

Here, we formalize the definition of generalization in relation to interpolation and extrapolation. We use a language and notation similar to the one used for definition of NP-hardness \citep{cormen2009introduction}.
%{jm do you have a ref for NP hardness?}

{\bf Notation:}

$\mathcal{G}:$ generalization,

$EP:$ extrapolation,

$IP:$ interpolation,

% $OD:$ out of distribution,

$\mathcal{M}:$ trained model/function,

$\Omega:$ domain of $\mathcal{M}$,

$\kappa:$ scope of the trained model/training set/testing set.

\vspace{1cm}

{\bf Definitions:}

Generalization in this context implies the ability of a model, $\mathcal{M}$, to accurately perform the classification task that it is trained for, on unseen data. Therefore, the model, the classification task/scope, and the unseen data are all related to each other. The model is trained to perform a certain classification task, and we expect it to perform the task on unseen data related to that task.

$\mathcal{M}$ accepts inputs, $x$, from a domain $\Omega$ (e.g., pixel space), and produces an output vector $$z = \mathcal{M}(x),$$ where $z$ has a constant number of elements, each corresponding to a certain class. The class (element of $z$) with the largest value is the classification of the model for a certain input. $\mathcal{M}$ is trained using a specific training set and that training set has a specific \hull.

Interpolation is when $\mathcal{M}$ is applied to any input inside the \hull 
\begin{equation} \label{eq:gen_inter}
	IP^{\mathcal{M}} := \{\mathcal{M}(x) , \forall x \in \mathcal{H}^{tr} \}.
\end{equation}
When an input to $\mathcal{M}$ is inside the \hull, we can say that $\mathcal{M}$ is interpolating between the its training samples when processing the input. Hence, interpolation is limited to the convex hull of training set. We consider \hull to be bounded since $\Omega$ is bounded and \hull is a subset of $\Omega$.% For brevity (similar to the notation used for NP-hardness), we can denote this by $$IP^{\mathcal{M}} \subset \mathcal{H}^{tr},$$ stating that interpolation is the set of classifications that $\mathcal{M}$ does for inputs inside its \hull.

Similarly, extrapolation is when $\mathcal{M}$ is used to classify any inputs outside the \hull 
\begin{equation} \label{eq:gen_extra}
EP^{\mathcal{M}} := \{\mathcal{M}(x) , \forall x \in  \Omega \backslash \mathcal{H}^{tr} \}.
\end{equation}
% alternatively, denoted by $$EP^{\mathcal{M}} \subset \Omega \backslash \mathcal{H}^{tr}.$$
If domain is bounded (e.g., $\Omega$ is a hyper-cube), then extrapolation would be over a bounded space.

Here, we are focused on how $\mathcal{M}$ processes an input as a function, and whether that process involves interpolation vs extrapolation. Our focus is not on the outputs produced by $\mathcal{M}$, rather, our focus is on how $\mathcal{M}$ processes its inputs in relation to its training samples, the samples used to derive the $\mathcal{M}$.%\footnote{So, our focus is not on $z$, but the act of classification performed by $\mathcal{M}$.}

By definition, interpolation and extrapolation do not have any overlap 
% \begin{equation*} \label{eq:gen2}
%     IP^{\mathcal{M}} \cap EP^{\mathcal{M}} = \emptyset,
% \end{equation*}
because 
\begin{equation} \label{eq:gen3}
    \mathcal{H}^{tr} \cap (\Omega \backslash \mathcal{H}^{tr}) = \emptyset.
\end{equation}
In other words, a particular input $x$ either belongs to \hull or not, and its relationship with respect to \hull will satisfy either equation~\eqref{eq:gen_inter} or~\eqref{eq:gen_extra}.

When we train a classification model, the model partitions its domain while partitions are defined by decision boundaries. Partitioning of the domain and location of decision boundaries inside the \hull can be explained in relation to the location of training samples. At the same time, decision boundaries may extend outside the \hull, throughout the $\Omega$. It is also possible that some isolated decision boundaries exist outside the \hull unrelated to the ones inside. Classification of inputs outside the \hull requires extrapolation, i.e., a model has to rely on the partitions and decision boundaries outside the \hull to classify such inputs.

{\bf Scope:} 

Scope of the model, $\kappa$, can be defined for a specific task, e.g., learning to classify handwritten digits, learning to classify chest X-rays. Generalization is defined in relation to the scope of the model. Scope, and by extension generalization, may entail only part of the domain. For example, domain of $\mathcal{M}$ may be a hyper-cube of 1,000 dimensions (images of 100 by 100 pixels), where pixel values vary between 0 and 1. Such domain could contain images of handwritten digits, images of objects, images of chest X-rays, radiology images, etc. If scope of the model is classifying handwritten digits, then it would not entail regions of the domain that contain other types of images. Parts of the domain that are unrelated to the scope of the model, inside and/or outside the \hull, will be considered out-of-distribution. In other words, scope of the model will define which parts of the domain are in-distribution and which parts are out-of-distribution. Therefore, we can divide the domain into two disjoint portions
$$\Omega^\kappa \cup \Omega^{\lnot \kappa} = \Omega \quad , \quad \Omega^\kappa \cap \Omega^{\lnot \kappa} = \emptyset,$$

where $\Omega^\kappa$ is the portion of domain related to the scope of the model $\mathcal{M}$, and $\Omega^{\lnot \kappa}$ is the portion of domain unrelated to the scope of $\mathcal{M}$. Domain of a model, $\Omega$, could be the entire pixel space, while $\Omega^\kappa$ would be the subset of domain that contains images relevant to the scope of the model. One may intend to use the model only on $\Omega^\kappa$ (e.g., radiology images of liver), and as a result, generalization would be defined over $\Omega^\kappa$, and not the entire pixel space.

{\bf Generalization:}

Generalization for $\mathcal{M}$ is tightly defined in relation to its scope, i.e., generalization is the set of classifications that $\mathcal{M}$ does for inputs related to its scope
\begin{equation}
 \mathcal{G}^{\mathcal{M}} := \{ \mathcal{M}(x), \forall x \in \Omega^\kappa \}.
\end{equation}

% For brevity, we denote this by $$\mathcal{G}^{\mathcal{M}} \subset \Omega^\kappa,$$

This division of the domain, between $\Omega^{\kappa}$ and $\Omega^{\lnot \kappa}$, is different from that of $\mathcal{H}^{tr}$ and $\Omega \backslash \mathcal{H}^{tr}$ since certain regions within the \hull may be unrelated to the scope of the model, and similarly, certain regions outside the \hull may be related to $\kappa$.\footnote{For example, there may be sub-regions inside a \hull where training set of a model does not have any samples, and such sub-regions may be alien to the model and possibly to its scope.}

% For interpolation, scope of the model may or may not entail all of the \hull. Similarly, for extrapolation, scope of the model may or may not entail all of the $\Omega \backslash \mathcal{H}^{tr}$. 

%In equation~\eqref{eq:gen1}, we recognized that generalization may have interpolation and extrapolation aspects. 

The model, $\mathcal{M}$, can be used to classify inputs inside and/or outside the \hull as long as inputs are inside $\Omega^{\kappa}$. As a result, generalization may entail interpolation and/or extrapolation
\begin{equation} \label{eq:gen1}
    \mathcal{G}^{\mathcal{M}} = \mathcal{G}_{IP}^{\mathcal{M}} \cup \mathcal{G}_{EP}^{\mathcal{M}},
\end{equation}
where $\mathcal{G}_{IP}^{\mathcal{M}}$ is generalization for $\mathcal{M}$ via interpolation, i.e., the set of classifications that $\mathcal{M}$ makes for inputs related to scope inside the hull
\begin{equation} \label{eq:gen_interG}
  \mathcal{G}_{IP}^{\mathcal{M}} := \{\mathcal{M}(x) , \forall x \in \Omega^\kappa \cap \mathcal{H}^{tr} \},
\end{equation}
% alternatively, denoted by $$\mathcal{G}_{IP}^{\mathcal{M}} \subset \Omega^\kappa \cap \mathcal{H}^{tr},$$

and $\mathcal{G}_{EP}^{\mathcal{M}}$ is generalization for $\mathcal{M}$ via extrapolation, i.e., the set of classifications that $\mathcal{M}$ makes for inputs related to scope, but outside the hull
\begin{equation} \label{eq:gen_extraG}
  \mathcal{G}_{EP}^\mathcal{M} := \{\mathcal{M}(x) , \forall x \in  \Omega^\kappa \backslash \mathcal{H}^{tr} \}.
\end{equation}
% alternatively, denoted by $$\mathcal{G}_{EP}^{\mathcal{M}} \subset \Omega^\kappa \backslash \mathcal{H}^{tr}.$$

This way, we can distinguish between the interpolation and extrapolation aspects of generalization while tying them to the model's scope (i.e., the subset of domain related to $\kappa$). Because of equation~\eqref{eq:gen3}, we can conclude that $\mathcal{G}_{IP}$ (generalization inside the \hull) and $\mathcal{G}_{EP}$ (generalization outside the \hull) do not have any overlap. Any testing sample, $x$, that $\mathcal{M}$ classifies, is either outside the \hull or not, and as a result either satisfies equation~\eqref{eq:gen_interG} or~\eqref{eq:gen_extraG}. When $x$ is outside the \hull, $\mathcal{M}$ has to extrapolate (i.e., rely on the partitions and decision boundaries outside the \hull) in order to classify it. Vice versa, when $x \in \mathcal{H}^{tr}$, its classification will be based on the partitions inside the \hull.
% $$\mathcal{G}_{IP} \cap \mathcal{G}_{IP} = \emptyset.$$

Distinguishing between $\mathcal{G}_{IP}$ and $\mathcal{G}_{EP}$ is useful as it would help us understand different components of generalization for a model. For example, a model may generalize well inside the \hull, but generalize poorly outside the \hull. One can quantify the generalization of a model, separately, inside and outside the \hull, simply by evaluating which testing samples are inside the \hull. This distinction between the two components of generalization can then be used to understand and diagnose the model. For example, to understand the extrapolation aspect of generalization, one may study the partitioning of the domain outside the \hull.

Distinguishing between $\Omega^\kappa$ and $\Omega^{\lnot \kappa}$ is useful as it ties the generalization of the model to its scope by excluding out-of-distribution images. Any classification that a model does for inputs in $\Omega^{\lnot \kappa}$ may not count towards its generalization. For example, performance of a model on radiology images of liver would be irrelevant, if the scope of the model is classifying handwritten digits.

Boundaries of \hull are well-defined, and one can compute exact points on those boundaries by solving a convex optimization problem \citep{yousefzadeh2021sketching}. However, identifying the exact boundaries that define the $\Omega_\kappa$ may be hard for image datasets. This is an open research problem, commonly known as out-of-distribution detection \citep{liang2017enhancing,hendrycks2016baseline,zhang2021understanding_ood}.

We may intend to use certain models exclusively for interpolation purposes, or for extrapolation purposes, or for a combination of both. This could further narrow down the definition of $\mathcal{G}$ for such models. However, in its general form, generalization may entail both interpolation and extrapolation as stated in equations~\eqref{eq:gen1}-\eqref{eq:gen_extraG}.

Finally, we note that a model may be a composition of distinct functions as is the case for deep neural networks. It may be insightful to study interpolation and extrapolation aspects of such sub-functions, separately, in relation to generalization. For example, one may consider one of the inner layers of a model and study the domain and range of that specific layer, the extrapolation and interpolation performed in that layer, and also the role of that layer in generalization. We performed such study for the last layer of a residual neural network in which the domain of the layer is the 64-dimensional feature space extracted by previous layers of the model, and its range is the output classes of the image classification model \citep{yousefzadeh2021hull}.

{\bf Scope matters:}
Deep learning and image classification is hardly the first field of study where the scope of a model has to be defined. In economics, there is a concept known as ``economy of scope" where expanding the scope of an economic activity leads to cost savings. Similarly, when ``economy of scale" is studied for economic activities, the underlying assumption is that the predefined ``scope" is kept constant. As we reviewed earlier, in cognitive science and psychology, extrapolation is studied in relation to learning, and in such studies, a specific scope is defined for extrapolation, and then that scope is studied. As the functional task of image classification models involve both interpolation and extrapolation, we have to study their extrapolation, as well as their interpolation, in the context of the specific scope defined for each model, the same scope that the model is intended to be used for. In image classification, the scope relates to the type of images the model is intended to classify.
% \jm{based on our discussion, what do you think about adding, "in image classification, the scope relates to the type of images the model is intended to classify"} \ry{Sounds great! Please go ahead and add it.}

% Extrapolation has similar meanings in philosophy, in literature, and in many other fields.

% \jm{I find the last sentence here a little confusing, can you comment further?}

% \ry{Sure, see if it makes better sense now.}
% \jm{It seems better, but when I read that sentence, I think used for what? The same scope that the model predictions from the model is intended to used in? Just want to clarify my understanding. also can you point me to a ref for the "economy of scope"}

% \ry{Tried to clarify that. Used for classifying handwritten digits, for example. Or used for classifying certain objects: cats, cars, frogs, ... as in CIFAR-10. One creates an image classification model for a specific task, i.e., to perform specific types of classification.}
% \jm{that makes sense. do you think the reader will be clear on that definition of scope? in images, would it be the type of images the model classifies?} \ry{Yes, it would be the type of images. Let me think about the best way to clarify it. That's a good point.}

% \ry{you can look it up here: \url{https://en.wikipedia.org/wiki/Economies_of_scope}}

\end{document}